%% file: main.tex
\documentclass[journal]{IEEEtran}
\usepackage{times}
\usepackage{soul}
\usepackage{url}
\usepackage[hidelinks]{hyperref}
\usepackage[utf8]{inputenc}
\usepackage[small]{caption}
\usepackage{graphicx}
\graphicspath{ {./fig/} }
\usepackage{amsmath}
\usepackage{amssymb}
\usepackage{amsfonts}
\usepackage{amsthm}
\usepackage{booktabs}
\usepackage{algorithm}
\usepackage{algorithmic}
\usepackage{stfloats}
\usepackage{multirow}
\usepackage{xspace}
\usepackage{hyperref}
\usepackage{setspace}
\usepackage{wrapfig}
\usepackage{cite}
\usepackage{authblk}
\usepackage[square,sort,comma,numbers]{natbib}

\newcommand{\etal}{\textit{et al}.\xspace}
\newcommand{\ie}{\textit{i}.\textit{e}.\xspace}
\newcommand{\eg}{\textit{e}.\textit{g}.\xspace}
\newcommand{\vct}[1]{\boldsymbol{#1}} 
\newcommand{\mat}[1]{\boldsymbol{#1}} 

\usepackage{dcolumn}
\newcolumntype{d}[1]{D{.}{.}{#1}}

\usepackage{xargs} 
\usepackage[pdftex,dvipsnames]{xcolor} 

\usepackage[colorinlistoftodos,textsize=scriptsize,textwidth=8mm]{todonotes}
\usepackage{marginnote}

\makeatletter
\let\save@mathaccent\mathaccent
\newcommand*\if@single[3]{%
  \setbox0\hbox{${\mathaccent"0362{#1}}^H$}%
  \setbox2\hbox{${\mathaccent"0362{\kern0pt#1}}^H$}%
  \ifdim\ht0=\ht2 #3\else #2\fi
  }
\newcommand*\rel@kern[1]{\kern#1\dimexpr\macc@kerna}
\newcommand*\widebar[1]{\@ifnextchar^{{\wide@bar{#1}{0}}}{\wide@bar{#1}{1}}}
\newcommand*\wide@bar[2]{\if@single{#1}{\wide@bar@{#1}{#2}{1}}{\wide@bar@{#1}{#2}{2}}}
\newcommand*\wide@bar@[3]{%
  \begingroup
  \def\mathaccent##1##2{%
    \let\mathaccent\save@mathaccent
    \if#32 \let\macc@nucleus\first@char \fi
    \setbox\z@\hbox{$\macc@style{\macc@nucleus}_{}$}%
    \setbox\tw@\hbox{$\macc@style{\macc@nucleus}{}_{}$}%
    \dimen@\wd\tw@
    \advance\dimen@-\wd\z@
    \divide\dimen@ 3
    \@tempdima\wd\tw@
    \advance\@tempdima-\scriptspace
    \divide\@tempdima 10
    \advance\dimen@-\@tempdima
    \ifdim\dimen@>\z@ \dimen@0pt\fi
    \rel@kern{0.6}\kern-\dimen@
    \if#31
      \overline{\rel@kern{-0.6}\kern\dimen@\macc@nucleus\rel@kern{0.4}\kern\dimen@}%
      \advance\dimen@0.4\dimexpr\macc@kerna
      \let\final@kern#2%
      \ifdim\dimen@<\z@ \let\final@kern1\fi
      \if\final@kern1 \kern-\dimen@\fi
    \else
      \overline{\rel@kern{-0.6}\kern\dimen@#1}%
    \fi
  }%
  \macc@depth\@ne
  \let\math@bgroup\@empty \let\math@egroup\macc@set@skewchar
  \mathsurround\z@ \frozen@everymath{\mathgroup\macc@group\relax}%
  \macc@set@skewchar\relax
  \let\mathaccentV\macc@nested@a
  \if#31
    \macc@nested@a\relax111{#1}%
  \else
    \def\gobble@till@marker##1\endmarker{}%
    \futurelet\first@char\gobble@till@marker#1\endmarker
    \ifcat\noexpand\first@char A\else
      \def\first@char{}%
    \fi
    \macc@nested@a\relax111{\first@char}%
  \fi
  \endgroup
}

\begin{document}

\title{PFA-GAN: Progressive Face Aging with Generative Adversarial Network}

\author{
\thanks{
Z. Huang, S. Chen, and J. Zhang are with Shanghai Key Lab of Intelligent Information Processing and the School of Computer Science, Fudan University, Shanghai 200433, China (email: zzhuang19@fudan.edu.cn; szchen19@fudan.edu.cn; jpzhang@fudan.edu.cn).}
\thanks{
H. Shan is with Institute of Science and Technology for Brain-inspired Intelligence and  MOE Frontiers Center for Brain Science, Fudan University, Shanghai 200433, China, and Shanghai Center for Brain Science and Brain-Inspired Technology, Shanghai 201210, China (email: hmshan@fudan.edu.cn).
}
\thanks{\textit{Corresponding author: Hongming Shan}}
Zhizhong Huang,
Shouzhen Chen,
Junping Zhang,
Hongming Shan
}

\maketitle

\input{abs}
\input{intro}
\input{relatedwork}
\input{method}
\input{exp}
\input{conc}


\input{appendix}

\IEEEpeerreviewmaketitle

\ifCLASSOPTIONcaptionsoff
  \newpage
\fi

\end{document}

%% file: abs.tex
\begin{abstract}
    Face aging is to render a given face to predict its future appearance, which plays an important role in the information forensics and security field as the appearance of the face typically varies with age. Although impressive results have been achieved with conditional generative adversarial networks (cGANs), the existing cGANs-based methods typically use a single network to learn various aging effects between any two different age groups. However, they cannot simultaneously meet three essential requirements of face aging---including image quality, aging accuracy, and identity preservation---and usually generate aged faces with strong ghost artifacts when the age gap becomes large. Inspired by the fact that faces gradually age over time, this paper proposes a novel progressive face aging framework based on generative adversarial network (PFA-GAN) to mitigate these issues. Unlike the existing cGANs-based methods, the proposed framework contains several sub-networks to mimic the face aging process from young to old, each of which only learns some specific aging effects between two adjacent age groups. The proposed framework can be trained in an end-to-end manner to eliminate accumulative artifacts and blurriness. Moreover, this paper introduces an age estimation loss to take into account the age distribution for an improved aging accuracy, and proposes to use the Pearson correlation coefficient as an evaluation metric measuring the aging smoothness for face aging methods. Extensively experimental results demonstrate superior performance over existing (c)GANs-based methods, including the state-of-the-art one, on two benchmarked datasets. The source code is available at~\url{https://github.com/Hzzone/PFA-GAN}.
\end{abstract}

\begin{IEEEkeywords}
  Face aging, generative adversarial networks, progressive neural networks, divide-and-conquer, image translation.
\end{IEEEkeywords}

%% file: intro.tex
\section{Introduction}

Face aging is to render a given young face to predict its future appearance with natural aging effects while preserving his/her personalized features. It has broad applications ranging from digital entertainment to information forensics and security; \eg, predicting the future appearances of lost children, and the cross-age face verification~\cite{ling2009face,park2010age,li2011discriminative,wu2012age}. Thanks to the appealing application value of face aging, the literature has proposed numerous methods to address this problem in the past two decades~\cite{lanitis2002toward,suo2010compositional}, especially the supervised-based deep neural networks explored in~\cite{Duong_2016_CVPR,wang2016recurrent,Duong_2017_ICCV}. However, these methods require massive paired faces of the same subject over a long period for training, which is impractical and cumbersome. 

To alleviate the need for paired faces, in recent years, generative adversarial networks (GANs)~\cite{goodfellow2014generative} and its variant, conditional GANs (cGANs)~\cite{mirza2014conditional}, have been widely used to train a face aging model with unpaired face aging data~\cite{antipov2017face,zhang2017age,wang2018face,yang2018learning,song2018dual,liu2019attribute,li2019global,li2019age,zhu2020look,sun2020facial}, achieving better aging performance than conventional methods such as physical model-based methods~\cite{suo2010compositional} and prototype-based methods~\cite{kemelmacher2014illumination}. The resultant methods can be roughly summarized into two categories: cGANs- and GANs-based methods. The cGANs-based methods~\cite{zhang2017age,wang2018face,li2019global,li2019age,zhu2020look,sun2020facial} typically train a single network to learn various aging effects between any two different age groups, with a target age group as the condition. However, the generated faces from these methods cannot simultaneously meet three important requirements for face aging: image quality, aging accuracy, and identity preservation. For example, the conditional adversarial autoencoder (CAAE)~\cite{zhang2017age} was proposed to learn a face manifold, traversing on which smooth age progression and regression can be achieved simultaneously. Its generated faces cannot preserve the identity well and are prone to blurriness. On the contrary, the GANs-based methods~\cite{yang2018learning,liu2019attribute} attempt to improve the performance from different perspectives. For instance, Yang~\etal designed a pyramid-architecture discriminator to estimate high-level age-related details~\cite{yang2018learning}, and Liu~\etal fed the facial attribute vectors into both the generator and discriminator to keep facial attributes consistent. However, these existing methods attempt to learn aging effects between any different age groups by a different network. That is, each network is trained independently. A major drawback is that they cannot guarantee a smooth aging result due to the intrinsic complexities of face aging, such as various expressions and races. The aged faces are sometimes even younger than these of previous age groups; see Appendix Fig.~\ref{fig:not_smooth_vis}. Besides, when the age gap between the source age and target one becomes large, these methods usually generate ghosted or blurry faces.

\begin{figure}
    \centering
    \includegraphics[width=1\linewidth]{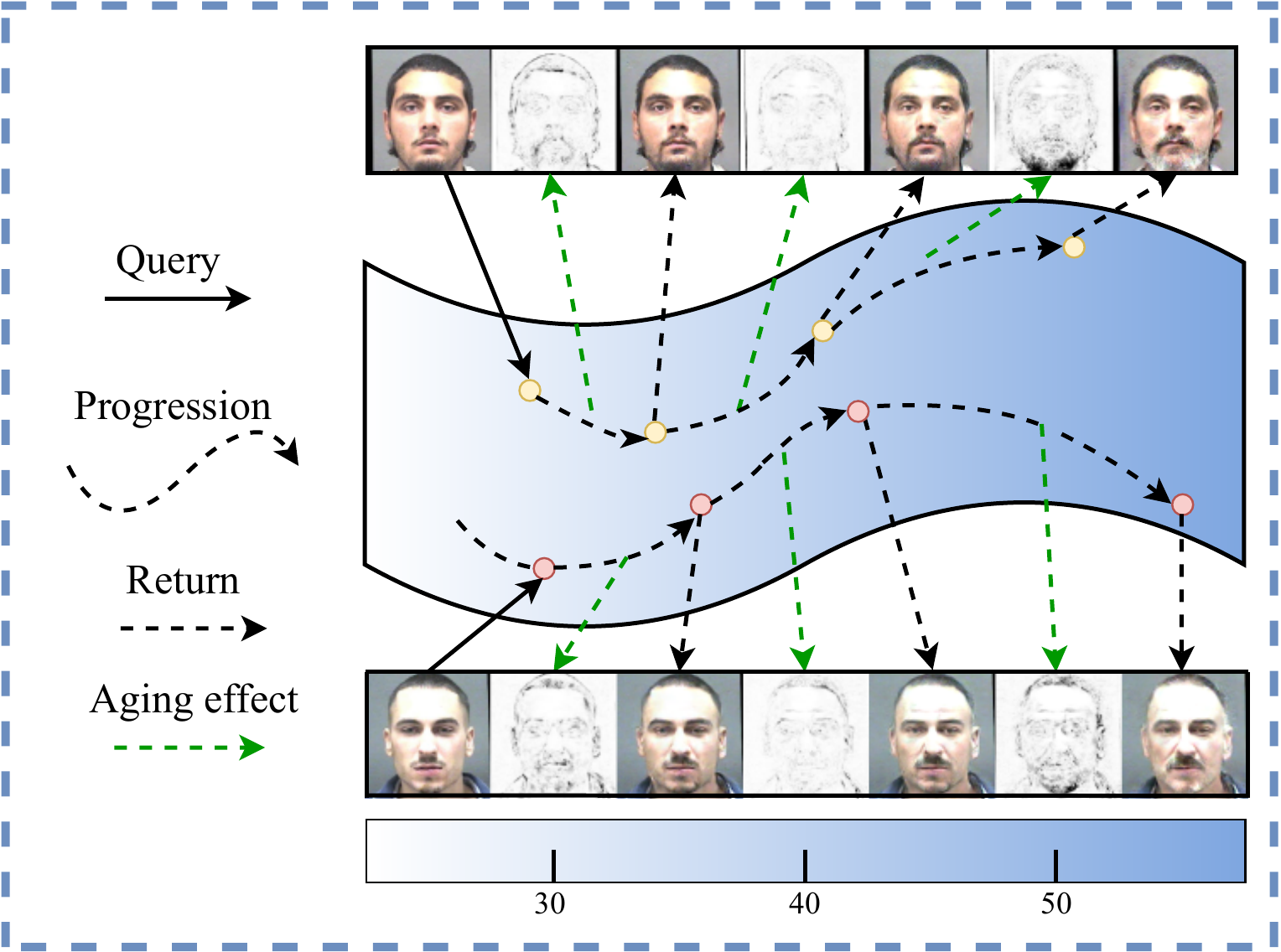}
    \caption{Illustration of face age progression on a face manifold. Faces traversing from light to dark appear different dominant aging effects at different age stages, which motivates us to model face aging in a progressive way.}
    \label{fig:mainfold}
\end{figure}

Inspired by the fact that faces gradually age over time, we are motivated to model the face aging process in a progressive way~\cite{karras2017progressive,shan2019competitive,zhang2019progressive}. Therefore, we propose a novel progressive face aging framework based on generative adversarial network~(PFA-GAN). More specifically, our PFA-GAN consists of multiple small sub-networks, each of which only deals with specific aging effects between two adjacent age groups. The rationale behind this idea is that the aging effects appear to be different at different ages; \eg, the phenomenon of hair turning white typically happens from 50 to 60 years old but is rare from 20 to 30 years old. To facilitate the understanding of this rationale, Fig.~\ref{fig:mainfold} depicts the face age progression on a face manifold, in which the dominant aging effects vary in the aging progress. In addition to that, the proposed framework can be trained in an end-to-end manner to alleviate the accumulative artifacts and blurriness, and generate smooth aging results. The main difference between PFA-GAN and previous GANs-based methods~\cite{yang2018learning,liu2019attribute} is that PFA-GAN trains several sub-networks \emph{simultaneously}, each of which aims at learning the aging effects between \emph{two adjacent} age groups, while previous GANs-based methods train several networks \emph{independently}, each of which learns the aging effects between \emph{any two} age groups. Experimental results on two large-scale age datasets empirically demonstrate the superiority of PFA-GAN over three state-of-the-art methods in generating high-quality faces with natural aging effects while preserving identity information. 

Here, we highlight the importance of modeling face aging in a progressive way in the following four aspects. First, progressive face aging focuses on modeling face aging effects between two adjacent age groups, which can decrease the learning complexity of face aging compared to traditional GAN-based or cGAN-based models that suffer from ghosted artifacts. Second, training progressive face aging in an end-to-end manner can enforce the network to produce smooth face aging results, which can further improve the image quality, aging accuracy, and identity preservation. Third, the ordinal relationship between age groups can be utilized to enhance the aging smoothness so that faces in a young group should be younger than the ones in an old group. Last, progressive face aging can improve the performance of cross-age verification, which is very important for security. As a by-product, progressive face aging can produce a sequence of aged faces for reference, facilitating cross-age verification in practice.

The main contributions of this paper are summarized as follows.
\begin{enumerate}
    \item[1)] We propose a progressive face aging framework based on generative adversarial network (PFA-GAN) to model the face age progression in a progressive way. 
    \item[2)] Unlike the traditional way that uses the target age group as a conditional input, we propose a novel age encoding scheme for PFA-GAN by adding binary gates to control the aging flow.
    \item[3)] We introduce an age estimation loss to take into account the age distribution for an improved aging accuracy.
    \item[4)] We propose to use the Pearson correlation coefficient (PCC) as an evaluation metric to measure the aging smoothness for face aging methods.
    \item[5)] We conduct extensive experiments on two benchmarked datasets to demonstrate the effectiveness and robustness of the proposed method in rendering accurate aging effects while preserving identity through both qualitative and quantitative comparisons.
\end{enumerate}

The rest of the paper is organized as follows. Sec.~\ref{related_work} surveys the development of face aging methods. In Sec.~\ref{sec:method}, we first formulate the face age progression into a progressive neural network, and then present the network architectures as well as the loss functions for PFA-GAN. This is followed by comprehensively comparing the proposed framework with recently published four state-of-the-art methods on two benchmarked age datasets in Sec.~\ref{sec:exp}. Finally, Sec.~\ref{sec:conc} presents a concluding summary.

%% file: relatedwork.tex
\section{Related Work}
\label{related_work}

Traditional methods of modeling face age progression can be roughly divided into two categories: physical model- and prototype-based methods. Physical model-based methods~\cite{lanitis2002toward,suo2010compositional} mechanically simulate the changes of facial appearance over time, such as muscles and facial skins, via a set of parameters. However, physical model-based methods are usually computationally expensive and do not generalize well due to the mechanical aging rules. On the contrary, prototype-based methods~\cite{rowland1995manipulating,suo2010compositional,kemelmacher2014illumination} compute the average faces of people in the same age group as the prototypes. As a result, the testing face can be aged by adding the differences between the prototypes of any two age groups. The main problem of prototype-based methods is that the personalized features cannot be preserved well due to the use of average faces.

Recently, deep neural networks have shown their potential for face aging~\cite{Duong_2016_CVPR,Duong_2017_ICCV,wang2016recurrent}. For example, Wang~\etal proposed a recurrent face aging framework that first maps the faces into eigenface subspace~\cite{turk1991face} and then utilizes a recurrent neural network to model the transformation patterns across different ages smoothly~\cite{wang2016recurrent}. As one of many supervised-based face aging methods, it requires massive paired faces of the same subject over a long period for training, which is impractical. On the contrary, PFA-GAN uses several sub-networks to learn aging effects with unpaired faces. 

To address this issue, many recent works utilize the availability of unpaired faces to train face aging models, either based on generative adversarial networks~(GANs)~\cite{goodfellow2014generative} or conditional GANs~(cGANs)~\cite{mirza2014conditional}. Most of unsupervised face aging methods are cGANs-based~\cite{antipov2017face,zhang2017age,wang2018face,song2018dual,li2019global,li2019age,zhu2020look,sun2020facial}, with a target age group as the condition. For example, Zhang~\etal proposed a conditional adversarial autoencoder~(CAAE)~\cite{zhang2017age} to achieve both age progression and regression simultaneously by traversing on a low-dimensional face manifold. Nevertheless, projecting faces into a latent subspace often compromises the image quality due to the reconstruction and fails to preserve the identity-related information~\cite{wang2016recurrent,zhang2017age}. 
To overcome the above deficiencies, Wang~\etal introduced an identity-preserved cGAN~(IPCGAN) that utilizes a perceptual loss based on a pre-trained model to preserve identity information~\cite{wang2018face}. Considering that face rejuvenation is the reverse process of face aging, Song~\etal extended dual GANs~\cite{yi2017dualgan} into dual conditional GANs~(Dual cGAN) framework to achieve both face age progression and regression~\cite{song2018dual}, and then Li~\etal combined this framework with a spatial attention mechanism to better keep the identity information and reduce ghost artifacts~\cite{li2019age}. 

There are a few works directly using GANs to model the aging progression between any two age groups~\cite{yang2018learning,liu2019attribute}. For example, Yang~\etal designed a discriminator with the pyramid architecture~(PAG-GAN)~\cite{yang2018learning}, which can estimate high-level age-related details through a pre-trained neural network. In addition to preserving identity information, Liu~\etal further found that the inconsistency of facial attributes still exists in previous works, and they fed the facial attribute vectors into both the generator and discriminator to suppress the unnatural changes of facial attributes~\cite{liu2019attribute}. 

In summary, cGANs-based methods typically learn different age mappings by one or two models with a target age group as the condition, while the GANs-based methods train several models for different age mappings separately. The difference between these two different kinds of methods turns out that cGANs-based methods are more flexible than GANs-based methods but GANs-based methods can produce better results in face aging. A significant difference from previous works is that our proposed PFA-GAN enjoys the best of both worlds; it is a GAN-based method but with a novel age encoding scheme, is more flexible than current (c)GAN-based methods, and most importantly, achieves the best performance in image quality, aging accuracy, and identity preservation. 

\begin{figure*}[t!]
    \centering
    \includegraphics[width=1.0\linewidth]{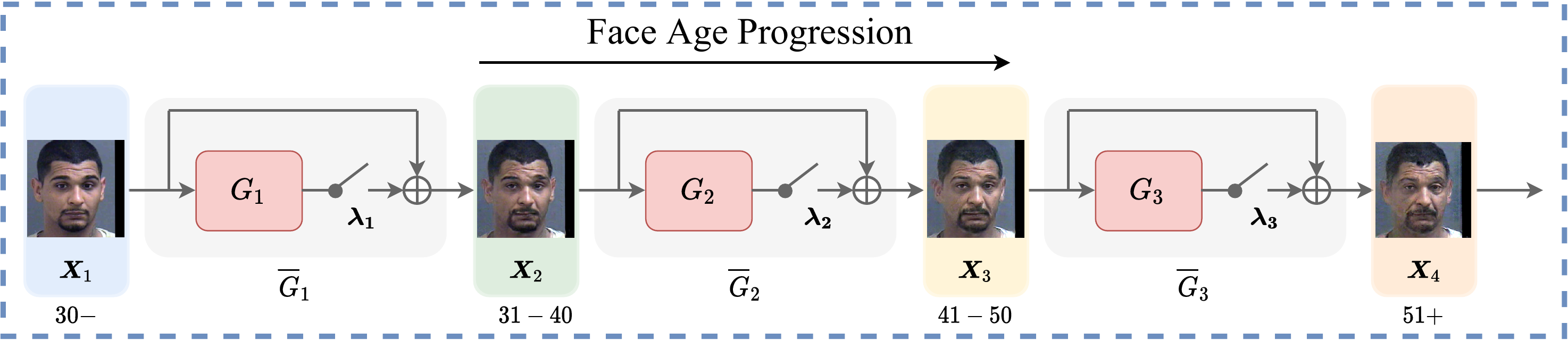}
    \caption{The proposed progressive face aging framework for a face aging task with 4 age groups. Each sub-network $\overline{G}_i$  aims at aging faces from age group $i$ to $i + 1$, which consists of a residual skip connection, a binary gate $\lambda_i$, and a light sub-network $G_i$ outputting aging effects.  The residual skip connection from input to output can prevent the sub-network from memorizing the exact copy of the input face.  The binary gate $\lambda_i$ can control the aging flow
    and determine if the aging mapping needs the sub-network $G_i$ to be involved.}
    \label{fig:framework}
\end{figure*}

%% file: method.tex
\section{Methodology}\label{sec:method}

Assuming that the faces lie on a face manifold residing in a high-dimensional space as shown in Fig.~\ref{fig:mainfold}, traversing from light to deep color achieves age progression. The change direction on the face manifold corresponds to the natural aging effects in the pixel space at different age stages. However, directly modeling the face manifold complicates the aging process of different races and sexes, resulting in low-quality faces and unexpected changes of identity. Therefore, we formulate this complicated aging process into a progressive neural network consisting of several sub-networks, each of which only learns the specific aging effects between two adjacent age groups in the image domain. The geodesic distance between any two age groups on the face manifold can be approximated by several locally linear Euclidean distances, similar to~\cite{tenenbaum2000global,he2018multi}.

Following~\cite{yang2018learning,liu2019attribute,li2019age}, we divide all ages into $N$ non-overlapping age groups, where $N=4$ in this paper. More specifically, the age ranges in age group $1$, $2$, $3$, and $4$ correspond to $30- $, $31 - 40$, $ 41 - 50$, and $51+ $, respectively. We interchange the age range and the age group index based on the context to simplify the expression without confusing. The following subsections present our novel progressive face aging framework, network architectures, and loss functions, respectively.

\subsection{Progressive Face Aging Framework}
\label{sec:PFA}

Prior to introducing our framework, we first briefly describe the cGAN-based methods. The cGANs-based methods employ one-hot encoding to represent the target age group $t$ as a vector $\vct{c}_{t} \in \mathbb{R}^{1\times N}$, whose elements are all $0$s except for a single $1$ to indicate the target age group. When given an input face image $\mat{X}_{s}\in \mathbb{R}^{w \times h \times 3}$ from source age group $s$, they first pad the vector $\vct{c}_{t}$ into a tensor $\vct{C}_t \in \mathbb{R}^{w\times h \times N}$, and then concatenate  $\mat{X}_s$ and $\mat{C}_t$ along the channel dimension as the input to an aging network $G$. Formally, the aged face $\mat{X}_t$ belonging to target age group $t$ can be expressed as:
\begin{align}\label{Eq:traditional}
    \mat{X}_t = G\Big([\mat{X}_s; \mat{C}_t]\Big),
\end{align}
where $[\mat{X}_s; \mat{C}_t]$ denotes the concatenation of two tensors $\mat{X}_s$ and $\mat{C}_t$ along the channel dimension. Although this encoding scheme is convenient to some extent, it enforces a single network to learn various aging effects between any two age groups, failing to generate high-quality faces when the age gap becomes large.

To address these issues, we propose a novel progressive aging framework inspired by the fact that faces gradually age over time. We formulate the aging process into a progressive neural network comprising several sub-networks, each of which only learns specific aging effects between two adjacent age groups. Fig.~\ref{fig:framework} shows that $i$-th sub-network $\widebar{G}_i$ is to age faces from age group $i$ to $i+1$; \ie, $\mat{X}_{i+1}=\widebar{G}_i(\mat{X}_i)$. Therefore, the progressive aging framework from the source age group $s$ to target one $t$ can be formulated as follows:
\begin{align}
    \mat{X}_t = \widebar{G}_{t-1}\circ \widebar{G}_{t-2}\circ\cdots\circ \widebar{G}_s (\mat{X}_s),
\end{align}
where symbol $\circ$ denotes the function composition. 

To prevent from memorizing the exact copy of the input face through several sub-networks, we employ a residual skip connection~\cite{he2016deep} to perform identity mapping from input to output in each sub-network, enabling sub-network to learn the aging effects. By introducing skip connection, we can easily recast the target age group into a sequence of binary gates that control the aging flow. The changes from age group $i$ to $i+1$ can be rewritten as:
\begin{align}
    \mat{X}_{i+1} =  \widebar{G}_i(\mat{X}_i) = \mat{X}_i + \lambda_i  G_i(\mat{X}_i).
\end{align}
To be clear, each sub-network consists of a residual skip connection, a binary gate, and the sub-network itself. Here, $\lambda_i \in \{0, 1\}$ is the binary gate controlling if the sub-network $G_i$ is involved in the path to target age group. That being said, $\lambda_i = 1$ if the sub-network $G_i$ is between source age group $s$ and target age group $t$, \ie, $s \leq i < t$; otherwise $\lambda_i = 0$. Put differently, we recast the tensor $\mat{C}$ used in the cGANs-based methods into a binary gate vector $\vct{\lambda}=(\lambda_1, \lambda_2, \ldots, \lambda_{N-1})$ controlling the aging flow for the proposed PFA-GAN framework.

With the proposed framework, the age progression, for example from age group $1$ to $4$ as shown in Fig.~\ref{fig:framework}, can be expressed as:
\begin{align}
    \mat{X}_4 & =  \mat{X}_3 + \underbrace{\lambda_3  G_3(\mat{X}_3)}_{\mathsf{aging\mbox{ }effects} :3\rightarrow 4} \\
              & =  \mat{X}_2 + \underbrace{\lambda_2  G_2(\mat{X}_2) + \lambda_3  G_3(\mat{X}_3)}_{\mathsf{aging\mbox{ }effects} :2\rightarrow 4} \notag\\
              & =  \mat{X}_1 + \underbrace{\lambda_1  G_1(\mat{X}_1) + \lambda_2  G_2(\mat{X}_2) + \lambda_3  G_3(\mat{X}_3)}_{\mathsf{aging\mbox{ }effects} :1\rightarrow 4}. \notag
\end{align}
When one wants to predict the aged face from age group $2$ to $3$, the above equation reduces to $\mat{X}_3 = \mat{X}_2 +  G_2(\mat{X}_2)$ since the gate vector for this aging mapping is $(0,1,0)$, which consequently bypasses sub-networks $G_1$ and $G_3$. It can be seen that the proposed framework is quite flexible in modeling the age progression between any two different age groups with this novel age encoding scheme.

Finally, given a young face $\mat{X}_{s}$ of source age group $s$, the aging process of $\mat{X}_{s}$ from $s$ into an old age group $t$ could be formulated as:
\begin{align}
    \mat{X}_{t} = G\Big(\mat{X}_{s}, \vct{\lambda}_{s:t}\Big),
\end{align}
where $G=\widebar{G}_{N-1}\circ \widebar{G}_{N-2}\circ\cdots\circ \widebar{G}_1$ denotes the entire progressive face aging network in PFA-GAN, and $\vct{\lambda}_{s:t}$ controls the aging process, where subscript $s\!:\!t$ indicates the elements in $\vct{\lambda}_{s:t}$ are all $0$s except for the indices from $s$ to $t-1$ that are $1$s. Note that PFA-GAN can also be applied to face rejuvenation by reversing the order of the age groups, which is detailed in Sec.~\ref{sec:exp}.

Considering that the (c)GAN-based models can also be used sequentially for face aging just like our PFA-GAN,  we call such variants of (c)GANs-based models sequential (c)GANs. Since the GANs and cGANs are used differently for face aging, we discuss the sequential (c)GANs separately. 
First, the cGANs-based methods typically learn different age mappings by one network with a target age group as the condition. Once trained, the cGANs can be used sequentially to age a given face by changing the condition accordingly. The disadvantages of the sequential cGANs can be summarized as follows. 1) Using one single network to learn various face aging effects between different age groups could produce ghosted artifacts, especially when the gap between age groups becomes large. 2) The intermediate generated faces are unseen to the network; the potential domain shift between generated and real faces for the latter cGANs could lead to inferior performance. 
Second, the GANs-based methods train several models for different age mappings separately. To be used sequentially, we consider the case that GANs-based methods train several sub-networks for adjacent age groups separately, which is similar to our proposed PFA-GAN. The drawback of the sequential GANs is that those sub-networks are trained separately, which cannot sense the generated faces for latter sub-networks. The artifacts produced by earlier sub-networks could be enlarged by the latter ones, leading to poor image quality. Different from the sequential (c)GANs, the unique advantage of PFA-GAN is that PFA-GAN optimizes these sub-networks in an end-to-end training, which can eliminate the accumulative errors and sense the generated faces from previous sub-network.

\subsection{Network Architecture}

\begin{figure}[t!]
    \centering
    \includegraphics[width=1.0\linewidth]{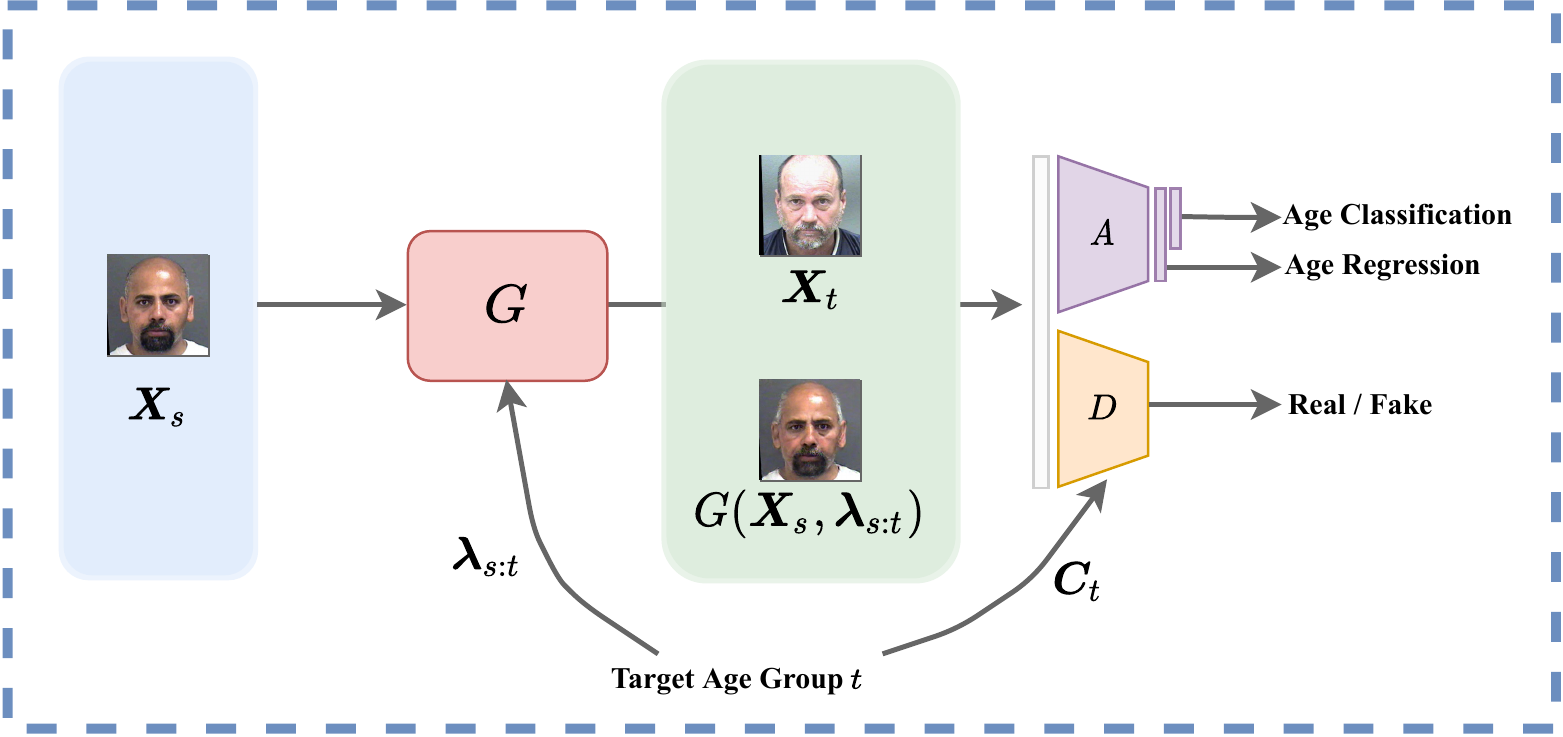}
    \caption{The GAN framework for PFA-GAN. The generator $G$ is to age a young face $\mat{X}_{s}$ into an aged face $G(\mat{X}_{s}, \vct{\lambda}_{s:t})$ that is indistinguishable from real face $\mat{X}_{t}$ by the discriminator $D$. $\vct{\lambda}_{s:t}$ is the binary gate vector to achieve progressive face aging from age group $s$ to $t$ for the generator while $\mat{C}_t$ is the age condition to align target age with the generated images for the discriminator. The pre-trained age estimation network $A$ is used to compute the age estimation loss---including an age group classification accuracy and an age regression error---for improved age accuracy and smoothness.}
    \label{fig:overall}
\end{figure}

As we described above, our PFA-GAN consists of several sub-networks to model the different age progression. Each sub-network only learns specific aging effects between two adjacent age groups. We use the GAN framework~\cite{goodfellow2014generative} to optimize each sub-network due to the lack of paired age dataset.  A GAN has two components: generator and discriminator. Without confusing, a sub-network is also called a sub-generator as the progressive face aging network serves as the generator in the GAN framework. Fig.~\ref{fig:overall} describes the GAN framework used for the proposed PFA-GAN, which comprises a generator $G$, a discriminator $D$, and an age estimation network $A$. The discriminator is to distinguish the fake aged faces from the real faces. The generator is to synthesize faces that fool the discriminator by producing more realistic faces. Each sub-generator should learn the aging patterns between two adjacent age groups. Therefore, compared to previous (c)GANs-based methods, our method can generate smooth aging results due to its nature of modeling the progressive aging. Here we describe the network architectures of the generator, discriminator, and age estimation network. The detailed network architectures can be found in Appendix Table~\ref{tab:network_architecture}.

\subsubsection{{Generator}}

When given $N$ age groups, our proposed PFA-GAN consists of $N-1$ sub-generators $\{\widebar{G}_i\}_{i=1}^{N-1}$. The input to each sub-generator is colorful facial images of size $256\times 256\times 3$. Similar to~\cite{zhu2017unpaired}, our network structure is a residual encoder-decoder network. The encoder has three convolutional layers that have respectively $32$ $9\times 9$, $64$ $4\times 4$, and $128$ $4\times 4$ convolution filters. Likewise, the decoder has two deconvolutional and one convolutional layers, whose filter sizes and the numbers of filters are the same as the encoder in reverse. We employ 4 residual blocks to convey the information from encoder to decoder. The skip connection from input to output enables the sub-network to learn aging effects without memorizing the exact copy of the input face.  All the convolutional layers except the last one are followed by instance normalization and leaky rectified linear unit (LReLU) activation whose slope is 0.2 for negative input. Unlike~\cite{liu2019attribute} that uses the tanh activation after the final layer to constrain the final output images within the range of $[-1,1]$, we found that the tanh activation is not appropriate in our framework, which limits the update of earlier sub-generators of $G$ during the training process. Instead, we did not use any activation function after the final layer.

\subsubsection{{Discriminator}}
\label{sec:discriminator}

We adopt the PatchDiscriminator from~\cite{isola2017image} as our discriminator $D$ thanks to its impressive results in a number of image-to-image translation tasks. The discriminator has a series of 6 convolutional layers with an increasing number of $4 \times 4$ filters. A spectral normalization layer~\cite{miyato2018spectral} and a LReLU activation with a slope of $0.2$ for negative input follow each convolutional layer except the first and last ones. Besides, similar to~\cite{wang2018face}, $\mat{C}_t$ is concatenated with the feature maps after the first convolutional layer for aligning conditions with the generated images.

\subsubsection{{Age Estimation Network}}
\label{sec:classifier}

To better characterize the face age distribution for an improved aging accuracy, an age estimation network $A$ with 6 convolutional layers and 1 fully-connected layer is incorporated into the progressive face aging framework. The last layer has 101 neurons corresponding to all possible ages from $0$ to $100$. Unlike~\cite{li2019age} that shares parts of parameters between the age estimation network and discriminator, we found that this sharing scheme could give rise to the training difficulty of GANs. Therefore, we separate them in PFA-GAN. It is important to note that the age estimation network has much fewer parameters than the pre-trained model adopted in~\cite{wang2018face,yang2018learning}.

Previous works typically employ either an age classification~\cite{wang2018face} term or an age regression~\cite{li2019age} term to check whether the generated face belongs to the target age group, which may be insufficient to characterize the face age distribution. In PFA-GAN, we adopt a deep expectation (DEX)~\cite{Rothe-IJCV-2018} term to learn the age distribution by computing a softmax expected value for age estimation. Formally, the estimated age $\widehat{\vct{y}}$ for an input face $\mat{X}$ can be computed as follows:
\begin{align}
    \widehat{\vct{y}}=\sum_{i=0}^{100} i \times \sigma\big[{A}(\mat{X})\big]_i,
\end{align}
where $\sigma$ represents a softmax function. The number of output neurons of the age estimation network ${A}(\mat{X})$ was empirically set to be 101 according to~\cite{Rothe-IJCV-2018}, where one neuron is expected to respond to one certain age. This should cover a wide age range from 0 to 100, and apparently includes the age ranges of the two datasets used in this paper. To regularize the learned age distribution for an improved learning efficiency, we train this age estimation network in a multi-task framework addressing the age regression and classification tasks simultaneously. We append another fully-connected layer with $N$ neurons on the top of ${A}$ for the age group classification task. By training the age estimation task in a multi-task framework, we found that the number of output neurons has no significant impact on the performance of the age estimation network as the age classification loss is able to regularize the learned age distribution. In addition, another consideration is that this setting could enable our framework to be adapted to other age estimation datasets such as the IMDB-WIKI~\cite{Rothe-IJCV-2018} and FG-NET~\cite{lanitis2002toward} with different age ranges.

Instead of training this age estimation network $A$ within our proposed framework, we found that it is better to pre-train this network on the training data. Once trained, the age estimation network is frozen and regularizes the generator for an improved aging accuracy.

\subsection{Loss Functions}

The overall loss functions used for PFA-GAN contain three components in order to meet these three requirements of face aging: 1) Adversarial loss aims to produce high-quality aged faces indistinguishable from real ones; 2) Age estimation loss expects to improve the aging accuracy; 3) Identity consistency loss seeks to preserve the same identity. These three loss components are detailed as follows. 

\subsubsection{{Adversarial Loss}}

The training of the GANs describes a competing game between a generator $G$ and a discriminator $D$, where $D$ aims to distinguish fake images from real ones, and $G$ attempts to fool $D$ by producing more realistic fake images. Once reaching a balance, $G$ can generate faithful images indistinguishable from the real ones. However, conventional GANs~\cite{goodfellow2014generative} suffer from maintaining a healthy competition between $G$ and $D$, leading to an unsatisfied performance. Therefore, we employ the least-squares GANs~\cite{mao2017least} for the discriminator to improve the quality of generated images and stabilize the training process. More specifically, least-squares GANs adopt the least-squares loss function to force the generator to generate samples toward the decision boundary rather than the negative log-likelihood loss function used in conventional GANs.

Given a young face $\mat{X}_{s}$ from age group $s$, the output of $G$ from $s$ to an old age group $t$ is $G(\mat{X}_{s}, \vct{\lambda}_{s:t})$. In the context of least-squares GANs, the adversarial loss for the generator $G$ is thus defined as:
\begin{align}
\mathcal{L}_{\mathrm{adv}}=\frac{1}{2} \mathbb{E}_{\mat{X}_{s}}\Big[D\big([G(\mat{X}_s, \vct{\lambda}_{s:t}); \mat{C}_t]\big)-1\Big]^{2}.
\end{align}

\subsubsection{{Age Estimation Loss}}

Apart from being photo-realistic, synthetic face images are also expected to satisfy the target age condition. Therefore, we include an age estimation network ${A}$ in our progressive face aging framework to regularize the face age distribution by minimizing age estimation loss---including age regression loss and age group classification loss.
The age estimation network $A$ is pre-trained on training data, and frozen in our framework, regularizing the generator towards more accurate age estimation. Formally, the age estimation loss between estimated age $\widehat{\vct{y}}$ and target age $\vct{y}$ for the generator $G$ is defined as:
\begin{align}
    \label{eq:age_estimation_loss}
    \mathcal{L}_{\mathrm{age}}=\mathbb{E}_{\mat{X}_{s}}\Big[\big\|\vct{y}-\widehat{\vct{y}}\|_2 + \ell\big({A}(\mat{X})\mat{W}, \vct{c}_t\big)\Big],
\end{align}
where $\mat{W}\in \mathbb{R}^{101\times N}$ denotes the final fully-connected layer for age group classification task outputting the age group logits, and $\ell$ is the cross-entropy loss for age group classification. Note that Eq.~\eqref{eq:age_estimation_loss} is also the loss function to pre-train age estimation network ${A}$ and $\mat{W}$.
 
\subsubsection{{Identity Consistency Loss}}

To preserve the identity-related information of the face and keep the identity-irrelevant information such as the background unchanged, we adopt a mixed identity consistency loss between the input face and generated one, which includes a pixel-wise loss, a structural similarity (SSIM) loss~\cite{wang2004image}, and a feature-level loss. These three losses for identity preservation are defined as follows:
\begin{align}
    \mathcal{L}_{\mathrm{pix}} &= \mathbb{E}_{\mat{X}_{s}}\Big|G(\mat{X}_{s}, \vct{\lambda}_{s:t})-\mat{X}_{s}\Big|, \\
     \mathcal{L}_{\mathrm{ssim}} &= \mathbb{E}_{\mat{X}_{s}}\Big[1-\mathrm{SSIM}\big(G(\mat{X}_{s}, \vct{\lambda}_{s:t}),\mat{X}_{s}\big)\Big], \\
      \mathcal{L}_{\mathrm{fea}}&=\mathbb{E}_{\mat{X}_{s}}\Big\|\phi(G(\mat{X}_{s}, \vct{\lambda}_{s:t}))-\phi(\mat{X}_{s})\Big\|^2_F.\label{Loss:IPL:fea}
\end{align}
The feature-level loss $\mathcal{L}_{\mathrm{fea}}$ is employed to keep identity consistency in a high-level feature space, where $\|\cdot\|_F$ represents the Frobenius norm. Here, $\phi$ in Eq.~\eqref{Loss:IPL:fea} denotes the activation output of the $10$th convolutional layer from the VGG-Face descriptor~\cite{parkhi2015deep}, which is adapted to extract the identity-related semantic representation of a face image. However, this loss alone could lead to unexpected changes to the identity-irrelevant information. Therefore, the image reconstruction loss, \ie, mean absolute error~(MAE) between the input and output images, is adopted so that the sparse aging effects are produced and the background is unchanged. It may produce some outliers in the resultant aging effects. We leverage the SSIM loss to balance the identity-related information for feature-level loss and identity-irrelevant information for MAE because SSIM loss can measure the local structure similarity and is a trade-off between MAE in the image space and feature-level loss in a high-level feature space.

Finally, the identity consistency loss for generator $G$ is defined as:
\begin{align}
    \mathcal{L}_{\mathrm{ide}}=(1 - \alpha_{\mathrm{ssim}})\mathcal{L}_{\mathrm{pix}} + \alpha_{\mathrm{ssim}} \mathcal{L}_{\mathrm{ssim}} + \alpha_{\mathrm{fea}} \mathcal{L}_{\mathrm{fea}},
\end{align}
where $\alpha_{\mathrm{ssim}}$ and $\alpha_{\mathrm{fea}}$ are the hyperparameters to control the balance between these three losses. 

\subsubsection{{Final Loss}}

To generate a photo-realistic face which belongs to the target age group and has the same identity as the input one, the final loss function to optimize generator $G$ is expressed as:
\begin{align}
    \mathcal{L}_{G}=\lambda_{\mathrm{adv}} \mathcal{L}_{\mathrm{adv}} + \lambda_{\mathrm{age}} \mathcal{L}_{\mathrm{age}} + \lambda_{\mathrm{ide}} \mathcal{L}_{\mathrm{ide}}.
\end{align}

The discriminator $D$ is optimized by minimizing the following loss:
\begin{align}
    \mathcal{L}_D =& \frac{1}{2} \mathbb{E}_{\mat{X}} \Big[D\big([\mat{X};\mat{C}]\big)-1\Big]^{2} +\notag\\
    &
    \frac{1}{2} \mathbb{E}_{\mat{X}_{s}} \Big[D\big([G(\mat{X}_{s},\vct{\lambda}_{s:t});\mat{C}_t]\big)\Big]^{2},
\end{align}
where the first term is computed over all real faces from all age groups and second term over all generated fake faces from all target age groups.

During the training phase, the generator $G$ and discriminator $D$ are updated alternatively until the training converges. By feeding the output of sub-generator $\widebar{G}_i$ as the input to the next sub-generator $\widebar{G}_{i+1}$, we can train our whole model in an end-to-end manner to eliminate the accumulative error. Note that there is only one discriminator in our PFA-GAN. We inject the age condition $\mat{C}_t$ into discriminator so that one discriminator can be adapted to different age groups with different conditions. Besides, the error from the latter sub-generator is backpropagated to former sub-generators. Because we train PFA-GAN in an end-to-end manner, consequently, there may exist leakage of aging effects between sub-generators. On the one hand, the aging effects between two adjacent age groups only reflect the primary pattern transformation, there are still some faces that cannot fit in this transformation and may be suitable for other two adjacent age groups. In this case, the leakage of aging effects between sub-generators is unavoidable. On the other hand, our end-to-end training could eliminate the accumulative error. For example, once one sub-generator amid the whole aging process produces ghosted faces, the following sub-generators can detect such anomalies and enforce that sub-generator to produce satisfied faces through back-propagation. In a sense, the latter sub-generators provide an attention mechanism for earlier ones to age faces effectively, which is more critical than the leakage of aging effects.

%% file: exp.tex
\section{Experiments}\label{sec:exp}

\begin{figure*}[t!]
    \centering
    \includegraphics[width=1.0\linewidth]{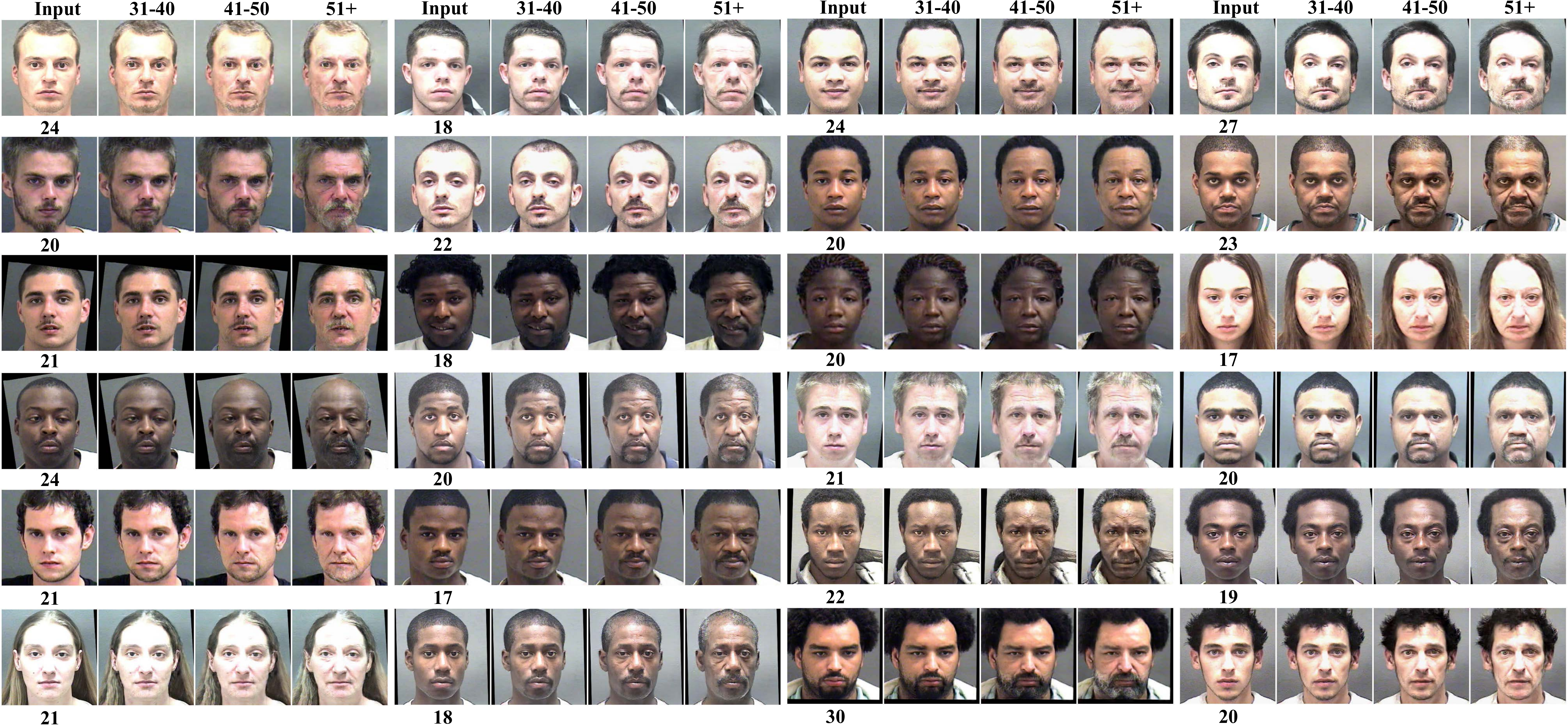}
    \caption{The generated aged faces by PFA-GAN on the MORPH dataset. We show the input faces with their real ages in the first column, and different aged faces of the same subject in the next three columns corresponding to three age groups $31-40$, $41-50$ and $51+$, respectively.}
    \label{fig:aged_faces_morph}
\end{figure*}

\subsection{Data Collection}

We conducted extensive experiments on two benchmarked age datasets: {MORPH}~\cite{ricanek2006morph}, and {CACD}~\cite{chen2015face}. The MORPH dataset~\cite{ricanek2006morph} is the most popular benchmark for face aging, which contains 55,134 colorful face images with strictly controlled conditions such as the near-frontal pose, neutral expression, moderate illumination, and simple background. The CACD dataset~\cite{chen2015face} contains 163,446 face images of 2,000 celebrities, which were collected via Google Image Search with few controlled conditions. Therefore, CACD has large variations in pose, illumination, and expression, making it much more challenging than the MORPH dataset.
For the raw face images in both datasets, we first extend the transformation matrix used in~\cite{zhang2016joint} to have an output size of $256\times 256$ with additional 20\% margin on all sides of the faces, doubling the image size of most previous works such as~\cite{zhang2017age,wang2018face}. We then align and crop the faces with five facial landmarks detected by Face++ API~\cite{faceplusplus.com} using an affine transformation to make the faces in the center of the input images. Before being fed into the network, the original intensity range of all cropped faces, $[0, 256]$, is linearly normalized into $[-1, 1]$. For the interpretation of the outputs, all the outputs of the network are first truncated into $[-1, 1]$ and then rescaled back to $[0,256]$ for the metrics calculation and display.
Following the convention~\cite{yang2018learning,liu2019attribute,li2019age}, we divided the face images into four age groups; \ie, $30-$, $31-40$, $41-50$, $51+$. For each dataset, we randomly select $80\%$ images for training and the rest for testing, and ensure that there is no overlap in identities between these two sets. Here, we also adopted FG-NET~\cite{lanitis2002toward} as testing set for a fair comparison with more prior works. Specifically, FG-NET is popular in face aging analysis but only contains 1,002 images from 82 individuals ranging from 0 to 69 and we used the model trained on CACD to age the faces from FG-NET.

\subsection{Implementation Details}

During training, we trained all models with a maximum of $200,000$ iterations on a single NVIDIA GTX 2080Ti GPU. PFA-GAN was initialized with He initialization~\cite{he2015delving}, and optimized by Adam optimization method~\cite{kingma2014adam}. The age estimation network was pre-trained on each dataset from scratch for $50$ epochs with a mini-batch size of $128$, an initial learning rate of $1.0\times10^{-4}$, and a learning rate decay factor of $0.7$ after every $15$ epochs. The generator and discriminator used the same training hyperparameters with learning rates of $1.0\times10^{-4}$, exponential decay rates for the first moment estimates $\beta_{1}$ of $0.5$, exponential decay rates for the second moment estimates $\beta_{2}$ of $0.99$, and mini-batch sizes of $12$. $G$ and $D$ were trained alternately. Input images were randomly sampled from the training set and normalized into the range of $[-1, 1]$. Note that in the testing phase, the final output images were clipped into the normal pixel range. The hyperparameters in the loss functions were empirically set as follows: $\lambda_{\mathrm{adv}}$ was $100$; $\lambda_{\mathrm{ide}}$ was $0.02$; $\lambda_{\mathrm{age}}$ was $0.4$; $\alpha_{\mathrm{ssim}}$ was $0.15$; and $\alpha_{\mathrm{fea}}$ was $0.025$. We implemented PFA-GAN based on PyTorch v1.3.1, and used Face++ APIs v3 to evaluate all methods.

\begin{figure*}[t!]
    \centering
    \includegraphics[width=1.0\linewidth]{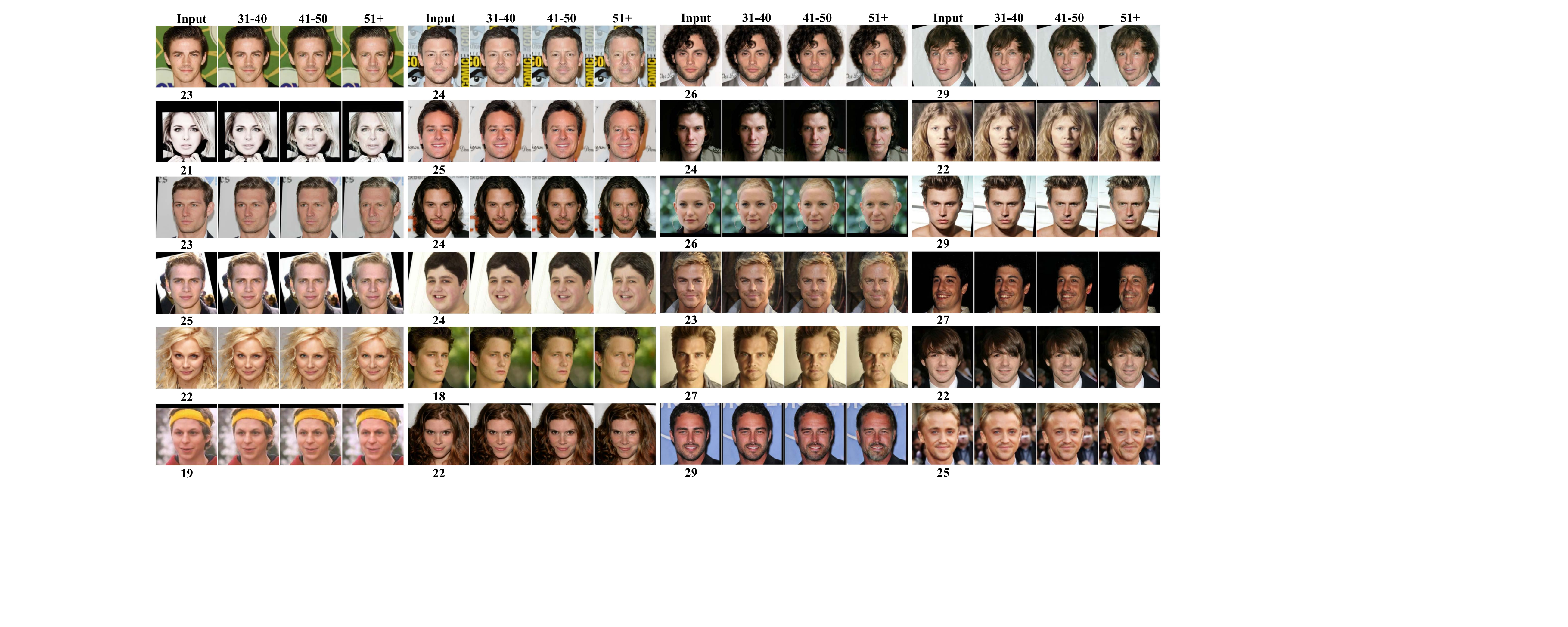}
    \caption{The generated aged faces by PFA-GAN on the CACD dataset. We showcase the input faces with their real ages in the first column, and different aged faces of the same subject in the next three columns corresponding to three age groups $31-40$, $41-50$ and $51+$, respectively.}
    \label{fig:aged_faces_cacd}
\end{figure*}

\begin{figure*}[t!]
    \centering
    \includegraphics[width=1.0\linewidth]{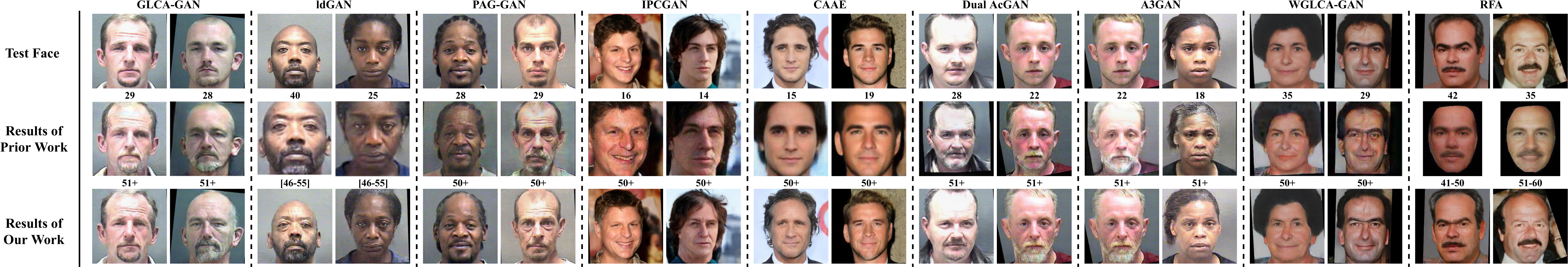}
    \caption{Performance comparison with prior work on the MORPH and CACD datasets. We showcase the input young faces in the first row with their real age labels below the image. The second row presents two sample results of prior work of seven recently published face aging methods. The third row shows our results of the same input faces in the same age groups as the prior work. Zoom in for a better view of image details.}
    \label{fig:method_comparison}
\end{figure*}

\subsection{Qualitative Evaluations}

Figs.~\ref{fig:aged_faces_morph} and~\ref{fig:aged_faces_cacd} showcase the aged faces from the MORPH and CACD datasets, respectively, which were generated by our PFA-GAN from the age group $30-$ to the other three old age groups. Although input faces cover a wide range of the population in terms of race, gender, pose, makeup, and expression, those aged faces are photo-realistic, with natural details in the skin, muscles, wrinkles, etc. For example, the beard turns into gray, and the skin gets wrinkles. Besides, all faces, even at a large age gap, can preserve their original identities. Although hair color generally turns white as the face ages, it varies from person to person and depends on the race and the training data, which explains why some generated faces in Figs.~\ref{fig:aged_faces_morph} and~\ref{fig:aged_faces_cacd} have few aging effects. Note that we have to compress the presented images for a reduced file size, which may cause some chessboard artifacts when zoomed in. 

\begin{figure*}[t!]
    \centering
    \includegraphics[width=1.0\linewidth]{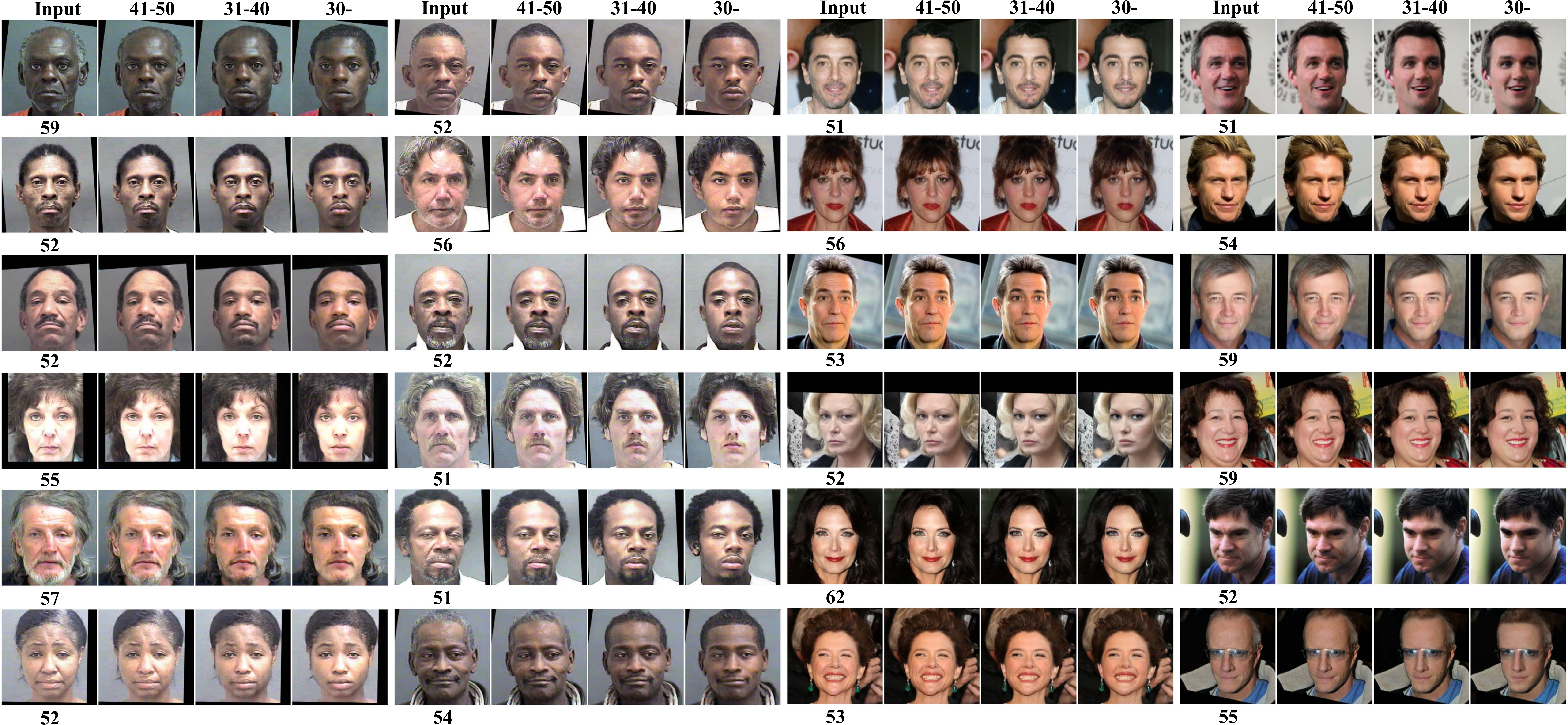}
    \caption{The rejuvenated faces by PFA-GAN on the MORPH (left two columns) and CACD (right two columns) datasets. We showcase the input faces with their real ages in the first column, and different rejuvenated faces of the same subject in the next three columns corresponding to three age groups $41-50$, $31-40$, and $30-$, respectively.}
    \label{fig:rejuvenated_faces}
\end{figure*}

To demonstrate the superiority of our progressive face aging framework, we compare the PFA-GAN with the most recently published state-of-the-art methods: GLCA-GAN~\cite{li2018global}, ldGAN~\cite{sun2020facial}, Pyramid-architectured GAN~(PAG-GAN)~\cite{yang2018learning}, IPCGAN~\cite{wang2018face}, CAAE~\cite{zhang2017age}, Dual AcGAN~\cite{li2019age}, Attribute-aware GAN~(A3GAN)~\cite{liu2019attribute}, WGLCA-GAN~\cite{li2019global}, and RFA~\cite{wang2016recurrent}. Note that we directly compared with the published results in their own papers, which is widely adopted in the face aging literature such as~\cite{yang2018learning,song2018dual,liu2019attribute,li2019global,li2019age,sun2020facial}. This can form a fair comparison for image quality and avoid any bias or error caused by our implementation. Fig.~\ref{fig:method_comparison} shows the comparison between PFA-GAN and the other baseline methods in generating high-quality aged faces. In contrast to previous cGANs-based methods including GLCA-GAN, ldGAN, IPCGAN, and CAAE, PFA-GAN achieves a better performance in aging faces of higher resolution ($2\times$) with enhanced aging details and realistic aging effects. Although PAG-GAN, Dual AcGAN, and A3GAN can generate high-quality face images, their aged faces usually contain ghost artifacts and unexpected changes of the background area. For instance, PAG-GAN fails to keep the background color where the clothes are also strongly ghosted. In addition, when the age gap becomes large, PFA-GAN is better at suppressing ghost artifacts and color distortion than Dual AcGAN and A3GAN. This benefits from the progressive aging framework, where the whole age progression is split into several small steps rather than a single step in the previous (c)GANs-based methods, and each sub-network only learns the adjacent aging translation patterns, which is relatively easy.

Although we mainly focus on face age progression, the proposed method can also be applied to face age regression. In the face rejuvenation, the input faces come from $51+$ age group, and are rejuvenated into three young age groups. Fig.~\ref{fig:rejuvenated_faces} shows the rejuvenated faces by our PFA-GAN. With the age decreasing from old to young, the face skin is tightened, and the hair becomes thick and luxuriant as expected. Moreover, PFA-GAN is not limited to face aging, it can also be easily extended to other image translation tasks; see Appendix Fig.~\ref{fig:expression_translation} for the sample results by applying PFA-GAN to smile-like facial expression translation.

\subsection{Quantitative Evaluations}

In this subsection, we used two widely-used and two auxiliary quantitative metrics to evaluate the performance of face aging methods: 
1) \textbf{Aging Accuracy} measures the difference of age distributions of both generic and synthetic faces in each age group; 
2) \textbf{Aging Smoothness} evaluates the capability of different methods in generating smooth aging results; 
3) \textbf{Inception Score} evaluates the image quality quantitatively; and 
4) \textbf{Identity Preservation} checks whether the identities have been preserved during face aging.

Following the convention~\cite{li2019age,liu2019attribute,yang2018learning}, we adopted the online face analysis tool developed by Face++ to estimate the face aging accuracy and identity preservation objectively. Moreover, we also conducted a double-blinded user study to evaluate the image quality of the generated faces subjectively. We compared the proposed model with previous state-of-the-art methods, including CAAE~\cite{zhang2017age}, IPCGAN~\cite{wang2018face}, WGLCA-GAN~\cite{li2019global}, and PAG-GAN~\cite{yang2018learning}, to demonstrate the effectiveness. Note that we have tried our best to reproduce the performance of these baseline methods as close as to that reported in their original papers. Specifically, the pre-trained AlexNet in IPCGAN and pre-trained LightCNN~\cite{wu2018light} in WGLCA-GAN were also replaced with VGG-Face descriptor for a fair comparison. We followed the original setting suggested in their original papers for CAAE and PAG-GAN.

\subsubsection{{Aging Accuracy}}

\begin{table*}[ht]
    \caption{Quantitative results of age estimation error of three age groups, the Pearson correlation coefficient (PCC), and the inception score (IS) on the MORPH and CACD datasets. For age estimation error, we report the absolute value between the mean estimate ages of real faces and fake faces in each age group. IS are calculated over all aged test faces. We also report the results of applying IPCGAN and PAG-GAN sequentially to age faces, which are denoted as IPCGAN$^\sharp$ and PAG-GAN$^\sharp$, respectively. The best results are highlighted in bold.}
    \label{tab:performance_age}
    \centering
    \begin{spacing}{1.0}
        \begin{tabular}{lcccccclccccc}
            \toprule
            \multicolumn{6}{c}{\textbf{MORPH}} & & \multicolumn{6}{c}{\textbf{CACD}} \\ 
            \cmidrule{1-6} \cmidrule{8-13} 
            \multirow{2}{*}{Method} & \multicolumn{3}{c}{Age Estimation Error} & \multirow{2}{*}{PCC} & \multirow{2}{*}{IS} && \multirow{2}{*}{Method} & \multicolumn{3}{c}{Age Estimation Error} & \multirow{2}{*}{PCC} & \multirow{2}{*}{IS}  \\ \cmidrule{2-4} \cmidrule{9-11} 
             & 31 - 40 & 41 - 50 & 51+ &  &  &&  & 31 - 40 & 41 - 50 & 51+ &  &  \\ 
            \midrule
            CAAE    & 4.41 & 5.84 & 6.07 & 0.937 & 2.38 && CAAE & 3.55 & 5.07 & 5.32 &  0.946 & \hspace{1.75mm}6.26\\
            IPCGAN  & 2.04 & 2.68 & 2.01 & 0.978 & 3.09 && IPCGAN & 1.78 & 0.50 & 2.64 & 0.972 & 29.07\\
			IPCGAN$^\sharp$ & 2.04 & 1.05 & 1.57 & 0.982 & 2.71 && IPCGAN$^\sharp$ & 1.78 & 3.00 & 1.66 & 0.978 & 22.58\\
            WGLCA-GAN & 3.52 & 1.41 & 2.98 & 0.974 & 3.15 && WGLCA-GAN & 0.63 & 1.75 & 2.33 & 0.981 & 29.60\\
            PAG-GAN & 0.77 & 0.43 & 1.56 & 0.955 & 3.67 && PAG-GAN & 0.94 & 1.09 & 1.27 & 0.951 & 26.43\\
            PAG-GAN$^\sharp$ & 0.77 & 1.01 & 1.34 & 0.968 & 2.87 && PAG-GAN$^\sharp$ & 0.94 & 1.12 & 0.72 & 0.971 & 19.20\\
            \cmidrule{1-6} \cmidrule{8-13}
            PFA-GAN & \textbf{0.38} & \textbf{0.14} & \textbf{1.11} & \textbf{0.989} & \textbf{3.90} && PFA-GAN & \textbf{0.41} & \textbf{0.11} & \textbf{0.37} & \textbf{0.986} & \textbf{33.39}\\
            \hspace{1.5mm}w/o DEX & 1.74 & 1.55 & 1.40 & 0.983 & 3.49 && \hspace{1.5mm}w/o DEX & 1.13 & 0.38 & 1.38 & 0.983 & 32.28\\
            \hspace{1.5mm}w/o PFA & 0.69 & 1.27 & 1.49 & 0.979 & 3.01 && \hspace{1.5mm}w/o PFA & 0.72 & 0.47 & 1.04 & 0.976 & 30.37\\
            \bottomrule
            \end{tabular}
    \end{spacing}
\end{table*}

In the mainstream works of face aging~\cite{yang2018learning,li2019global,liu2019attribute,li2019age,sun2020facial,zhu2020look}, the discrepancy between the age distributions of both generic and synthetic faces in each age group, also referred to as {age estimation error}, is a widely-used evaluation metric to measure the aging accuracy of different face aging methods. Specifically, the ages of both real and fake faces in each age group are first estimated by Face++ APIs for a fair comparison, and then the discrepancy between the mean ages of real and fake faces from the same age group is the age estimation error, where a lower value indicates a more accurate simulation of aging effects. Following the convention, only young faces from the age group of $30-$ are considered as the testing samples, and their aged faces in the other three age groups are produced by different methods. Table~\ref{tab:performance_age} presents the age estimation errors of different methods for each age group on the MORPH and CACD datasets. Markedly, our PFA-GAN consistently outperforms other baseline methods by a large margin in these three age groups, even when the age gap becomes large. CAAE fails to produce strong enough aging effects so that the synthetic faces are also over-smoothed with subtle changes, leading to large errors in estimated ages. Compared to IPCGAN, PAG-GAN performs better in aging faces with enhanced details due to the pyramid architecture. Modeling face age progression in a progressive way renders PFA-GAN achieve the best performance in aging accuracy among all methods, in which the difficulties of learning several aging translation patterns in cGANs-based methods are significantly reduced.

\subsubsection{{Aging Smoothness}}

Although aging accuracy, or age estimation error, can evaluate the performance of the aging accuracy of different methods, this metric cannot reflect the aging smoothness of individual faces, which is another important metric for face aging methods. When achieving a satisfied aging accuracy, face aging models should, as expected, produce a smooth aging process. Intuitively, given the ages of input faces and aged faces belonging to $N-1$ old age groups, their relative positions to the mean ages in the age distribution should be the same for all age groups. Therefore, a linear correlation exists between the age sequence of one subject and the mean age of generic one. To quantitatively calculate this linear correlation, we propose to use the Pearson correlation coefficient~(PCC) as a novel metric to measure the aging smoothness of different methods. Specifically, PCC has an output value between $-1$ and $+1$, where $1$ is a total positive linear correlation, $0$  no linear correlation, and $-1$ total negative linear correlation. We further calculate the mean PCC on all testing samples, which is defined as follows:
\begin{align}
    \mathrm{PCC} = \frac{1}{m}\sum_{i=1}^{m} \rho(\mathcal{Y}_i, \widebar{\mathcal{Y}})
\end{align}
where $\rho$ is the Pearson correlation coefficient function, $m$ the total number of samples, $\widebar{\mathcal{Y}}$ the generic age sequence including the mean ages of each age group from real data, and $\mathcal{Y}_i$ the $i$-th age sequence containing $N$ ages of the input face and the other $N-1$ aged faces. Note that the face ages were estimated by Face++ APIs~\cite{faceplusplus.com}. The higher PCC indicates not only a more smooth aging result but also a higher aging accuracy. Table~\ref{tab:performance_age} shows that CAAE fails to produce a smooth aging result with a lower aging accuracy. Although PAG-GAN performs better than IPCGAN in aging accuracy, IPCGAN could generate more smooth faces. This is because PAG-GAN learns each mapping separately and ignores the age distribution of whole training data. In contrast, with the progressive face aging framework, PFA-GAN learns each adjacent age mapping in an end-to-end manner and thus achieves the best performance in aging smoothness.

\subsubsection{{Inception Score}}

As a standard metric in image generation and translation, inception score (IS)~\cite{salimans2016improved} is widely used to evaluate the quality and diversity of generated images. Following~\cite{wang2018face} that does not suggest using the inception score based on a pre-trained network on ImageNet to evaluate the image quality of faces, we use the OpenAI source code to compute the inception score based on the pre-trained VGG-Face descriptor~\cite{parkhi2015deep}. Table~\ref{tab:performance_age} presents the inception scores for all methods. Due to the limited variation in the MORPH dataset, the inception scores are much lower than the ones on the CACD dataset. Note that all results are calculated over aged test faces for a fair comparison, hence the results of IPCGAN are not comparable to the one reported in the original paper. Obviously, PFA-GAN achieved the best results in terms of the inception score, indicating that PFA-GAN tends to produce higher image-quality faces than others.

\subsubsection{{Identity Preservation}}

\begin{table*}[ht]
    \caption{Quantitative results of face verification confidence and rate on the MORPH and CACD datasets. For the face verification rate, the false accept rate~(FAR) and threshold in Face++ APIs were set to be $10^{-5}$ and $76.5$, respectively. We also report the results of applying IPCGAN and PAG-GAN sequentially to age faces, which are denoted as IPCGAN$^\sharp$ and PAG-GAN$^\sharp$, respectively. The best results are highlighted in bold.}
    \label{tab:performance_id}
    \centering
    \begin{spacing}{1.0}
        \begin{tabular}{lcccclccc}
            \toprule
            \multicolumn{4}{c}{\textbf{MORPH}} & & \multicolumn{4}{c}{\textbf{CACD}} \\ 
            \cmidrule{1-4} \cmidrule{6-9} 
            Age Group & 31 - 40 & 41 - 50 & 51+ && Age group & 31 - 40 & 41 - 50 & 51+ \\ 
            \midrule
            \multicolumn{4}{c}{Verification Confidence} & & \multicolumn{4}{c}{Verification Confidence} \\ 
            \cmidrule{1-4} \cmidrule{6-9} 
            30 -    & 95.36 & 93.23 & 88.28 && 30 -    & 96.21 & 94.50 & 89.80 \\ 
            31 - 40 & --    & 96.20 & 92.87 && 31 - 40 & --    & 95.52 & 91.70 \\ 
            41 - 50 & --    & --    & 95.47 && 41 - 50 & --    & --    & 94.59 \\ 
            \multicolumn{4}{c}{Verification Rate~(\%)} & & \multicolumn{4}{c}{Verification Rate~(\%)} \\ 
            \cmidrule{1-4} \cmidrule{6-9} 
            CAAE    & \hspace{1.75mm}44.02   & \hspace{1.75mm}27.87   & \hspace{1.75mm}5.97 && CAAE    & 13.59 & \hspace{1.75mm}8.75 & \hspace{1.75mm}2.67\\
            IPCGAN  & \textbf{100.00}   & \textbf{100.00}   & 99.21 && IPCGAN  & 99.76 & 99.72 & 99.07\\
            
            IPCGAN$^\sharp$ & \textbf{100.00} & \hspace{1.75mm}99.92 & 77.12 && IPCGAN$^\sharp$ & 99.76 & 99.01 & 96.34\\

            WGLCA-GAN & \textbf{100.00} & \textbf{100.00}  & 98.82 && WGLCA-GAN & 99.90 & 99.88 & 98.89\\
            PAG-GAN & \textbf{100.00} & \hspace{1.75mm}98.97  & 91.51 && PAG-GAN & 99.93 & 99.38 & 97.87\\

            PAG-GAN$^\sharp$ & \textbf{100.00} & \hspace{1.75mm}93.69 & 59.16 && PAG-GAN$^\sharp$ & 99.93 & 98.01 & 89.24\\
            \cmidrule{1-4} \cmidrule{6-9}
            PFA-GAN & \textbf{100.00} & \textbf{100.00} & \textbf{99.70} && PFA-GAN & \textbf{99.97} & \textbf{99.89} & \textbf{99.69} \\
            \hspace{1.5mm}w/o DEX & \textbf{100.00}   & \textbf{100.00}   & 99.44 && \hspace{1.5mm}w/o DEX & 99.89 & 99.80 & 99.44\\
            \hspace{1.5mm}w/o PFA & \textbf{100.00}   & \textbf{100.00}   & 99.32 && \hspace{1.5mm}w/o PFA & 99.95 & 99.85 & 99.37\\
            \bottomrule
            \end{tabular}
    \end{spacing}
    
\end{table*}

Face verification experiments are conducted with Face++ APIs to examine identity preservation during face age progression. For each input young face, we checked if the aged and the input faces have the same identities. Following~\cite{yang2018learning,liu2019attribute}, we report the verification confidence between the synthetic aged images from different age groups of the same subject in the top portion of Table~\ref{tab:performance_id}, where the high verification confidence demonstrates that the identity information is consistently preserved. We also report the results of face verification rate in the bottom portion of Table~\ref{tab:performance_id}, in which the false accept rate (FAR) and threshold used in Face++ APIs were set to be $10^{-5}$ and $76.5$, respectively, as suggested by~\cite{yang2018learning,liu2019attribute}. With the great improvement of face aging accuracy on these two benchmarked datasets, PFA-GAN achieves the highest verification rate on all three age groups, and outperforms previous state-of-the-art methods by a large margin, especially in the challenging case from $30-$ to $51+$. CAAE fails to preserve the identity permanence since it maps the faces into a latent vector. Compared to IPCGAN that outperforms PAG-GAN thanks to the identity-preserved module, PFA-GAN still performs better in both aging accuracy and identity preservation. The main reason is that IPCGAN re-estimates all pixels in the output images, leading to a drop in performance, especially for in-the-wild images in the CACD dataset. Benefiting from the progressive face aging framework, each sub-network learns a residual image---aging effects---which greatly improves the identity preservation of PFA-GAN. Note that it is reasonable that the verification rate and confidence slightly decrease as more changes appear in the face when the age gap becomes age. Therefore, face verification results show that our method was relatively robust to preserve the identity information of input faces regardless of various attributes such as races and sexes.

\subsubsection{{Double Blinded User Study}}

We further conducted a double-blinded user study to evaluate all generated faces quantitatively from the perspective of human beings on Amazon Mechanical Turk~(AMT). First, we randomly selected $20$ real young faces from each dataset and collected their generated faces by the baseline methods as well as our method for the age group of $51+$. Second, $50$ volunteers evaluated all the aged faces with the reference input real young faces. These aged faces were in a randomly shuffled order. Finally, we asked them to select which one is the most realistic aged face at the age of over 51 years old, which should be 1) the same identity as the input face, 2) with natural aging effects, and 3) without ghost artifacts.

The odds of the aged face being selected as the most realistic one are $6.6\%$ for CAAE, $19.0$\% for IPCGAN, $30.0\%$ for PAG-GAN, and $44.4\%$ for PFA-GAN, respectively. Among all methods, our PFA-GAN achieves the best performance in this double-blinded user study. CAAE fails to produce photo-realistic faces with the identities preserved while IPCGAN produced lower resolution images~($2\times$) with blurry aging effects in contrast to PAG-GAN and PFA-GAN. Additionally, PAG-GAN fails to maintain identity consistency as good as PFA-GAN, making the results worse than PFA-GAN. In summary, our PFA-GAN can not only generate visually photo-realistic faces but also outperform other methods in aging accuracy and identity preservation.

\subsection{Comparison with Sequential (c)GANs}

\begin{figure}[t!]
    \centering
    \includegraphics[width=1.0\linewidth]{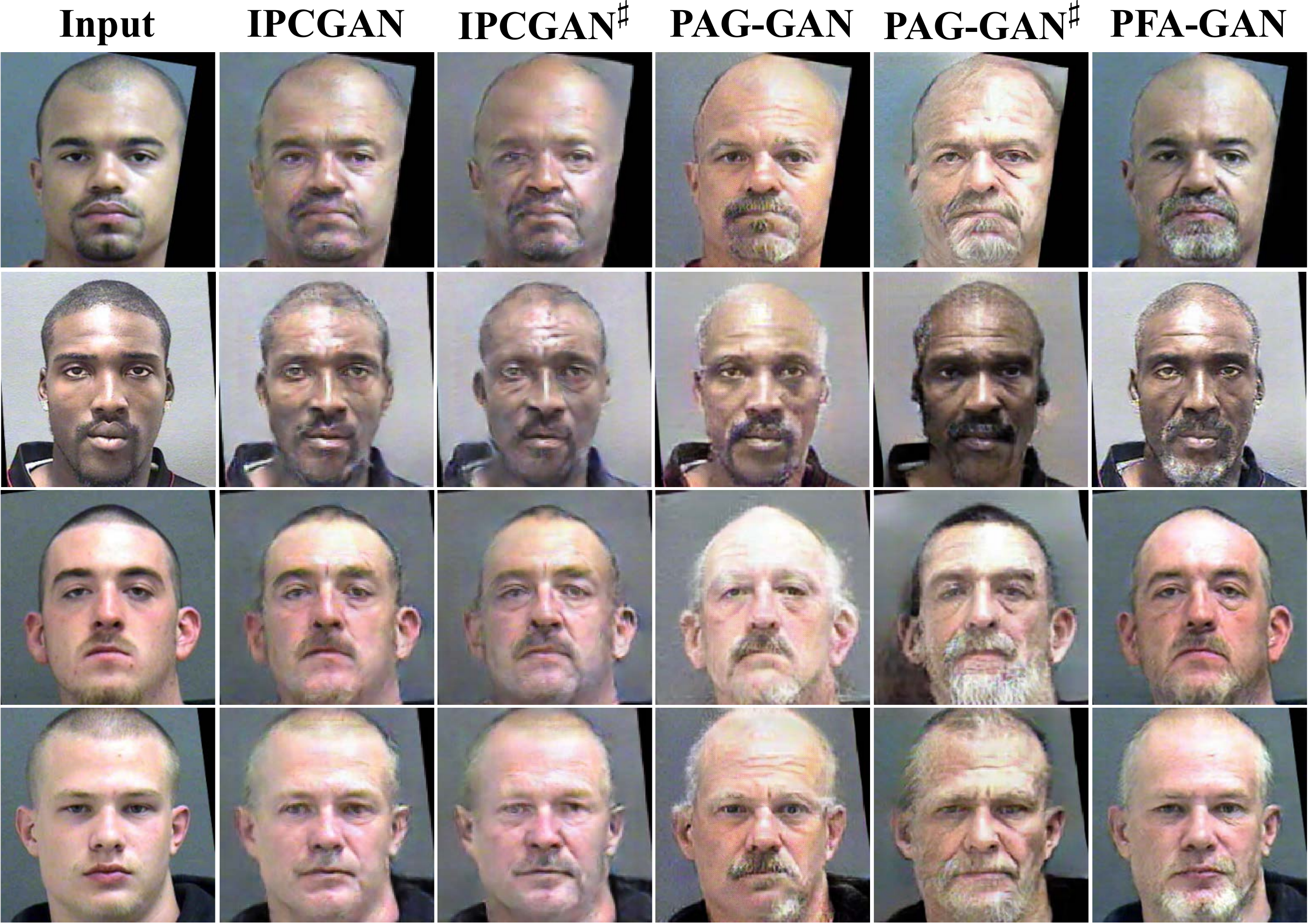}
    \caption{
        The generated aged faces of different methods to investigate the impact of sequential use on the (c)GANs-based methods including IPCGAN~(cGANs-based method) and PAG-GAN~(GANs-based method), which are denoted as IPCGAN$^\sharp$ and PAG-GAN$^\sharp$, respectively. The input faces from the age group of $30-$ are aged into the age group of $51+$.
    }
    \label{fig:sequential_vis}
\end{figure}

As discussed in~Sec.~\ref{sec:PFA}, the advantage of our PFA-GAN over the sequential (c)GANs is the end-to-end training. To demonstrate the difference between our PFA-GAN and sequential (c)GANs through both quantitative and qualitative comparisons, we implemented the sequential IPCGAN and sequential PAG-GAN as IPCGAN$^\sharp$ and PAG-GAN$^\sharp$, respectively, and reported their results in Fig.~\ref{fig:sequential_vis}, Tables~\ref{tab:performance_age} and~\ref{tab:performance_id}. Fig.~\ref{fig:sequential_vis} shows that when (c)GANs-based methods are applied sequentially to age faces, the aged faces are strongly ghosted and blurry with unsatisfied image quality. Quantitatively, the inception scores in Table~\ref{tab:performance_age} also drop as expected. It should be noted that the sequential use could improve the aging accuracy for some age groups, and  improve the aging smoothness at the cost of compromising identity preservation as shown in Table~\ref{tab:performance_id}, especially when the age gap becomes large.

However, different from sequential (c)GANs, our PFA-GAN trains these sub-networks simultaneously in an end-to-end manner. Consequently, PFA-GAN are able to remove accumulative error and produce satisfied faces through back-propagation. Besides, benfitting from the end-to-end manner, PFA-GAN provides the latter sub-networks with the capability of sensing the generated faces to eliminate the potential domain shift, and only focusing on changing these area of the face image relevant to face aging. As a result, PFA-GAN could not only further improve the image quality and aging accuracy with smooth aging results, but also maintain the identity consistency better than sequential (c)GANs. 

\subsection{Ablation Study}
\label{sec:ablation}

\begin{figure}[t!]
    \centering
    \includegraphics[width=1.0\linewidth]{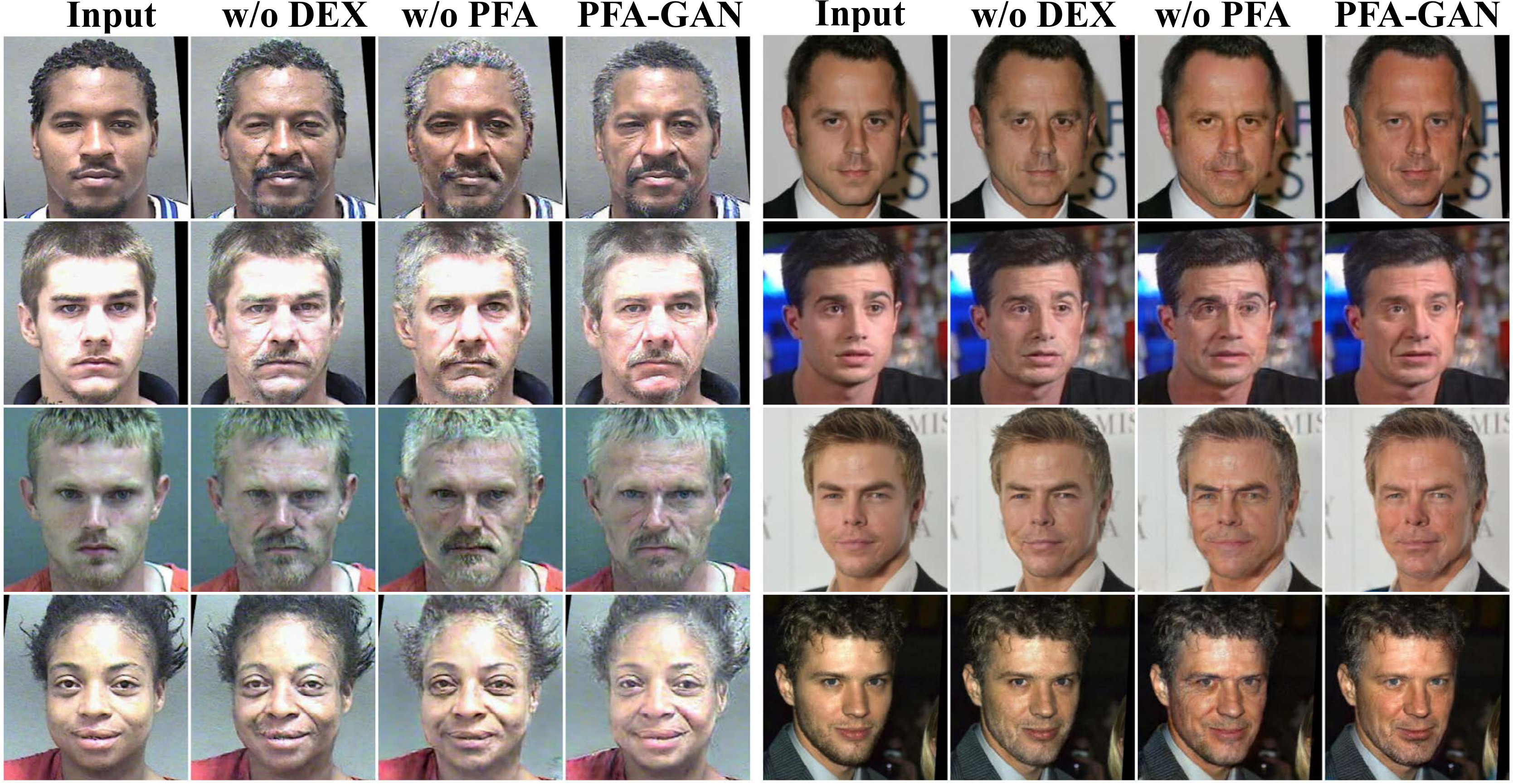}
    \caption{The ablation study results of PFA-GAN. The input faces from the age group of $30-$ are aged into the age group of $51+$ by PFA-GAN without DEX (w/o DEX), PFA-GAN without PFA (w/o PFA), and PFA-GAN itself.}
    \label{fig:ablation_study}
\end{figure}

In this subsection, an ablation study of PFA-GAN is conducted to fully explore the importance of DEX term and progressive face aging framework~(PFA) in simulating accurate age translations. We investigate the impact of these two modules by removing DEX term in loss function~(w/o DEX) and training only one sub-network that uses $\mat{C}_t$ to control the aging process like cGANs~(w/o PFA). When replacing DEX term with an age group classification loss, we increased $\lambda_{\mathrm{age}}$ from $0.4$ to $8$ for a fair comparison. Therefore, without DEX term, PFA-GAN only optimizes a single age group classification task in age estimation loss, and without progressive face aging framework, PFA-GAN reduces to a common cGANs-based method.

Fig.~\ref{fig:ablation_study} shows the visual comparison of face images generated by different variants of the proposed model. Highlighted by the hair and beard of the generated faces, PFA-GAN without DEX term fails to produce enhanced aging effects, and PFA-GAN without PFA suffers from severe ghost artifacts. On the contrary, the integration of DEX and PFA suppresses the ghost artifacts and produce realistic aging effects. To be specific, DEX is used to achieve an improved aging accuracy and PFA divides the whole aging process into several small steps towards a better image quality especially when the age gap becomes large.

Tables~\ref{tab:performance_age} and~\ref{tab:performance_id} show the quantitative results for ablation study. The results in Table~\ref{tab:performance_age} indicate that although introducing DEX greatly improves the aging accuracy, it fails to generate smooth aging results as the PFA does. Besides, the progressive face aging framework can achieve a better image quality than cGANs-based methods. With the improved image quality and aging accuracy, PFA-GAN achieves the best face verification rate in Table~\ref{tab:performance_id} since the identity-related features are preserved by the identity consistency loss.

\subsection{Robustness Analysis}

\begin{figure}[t!]
    \centering
    \includegraphics[width=1.0\linewidth]{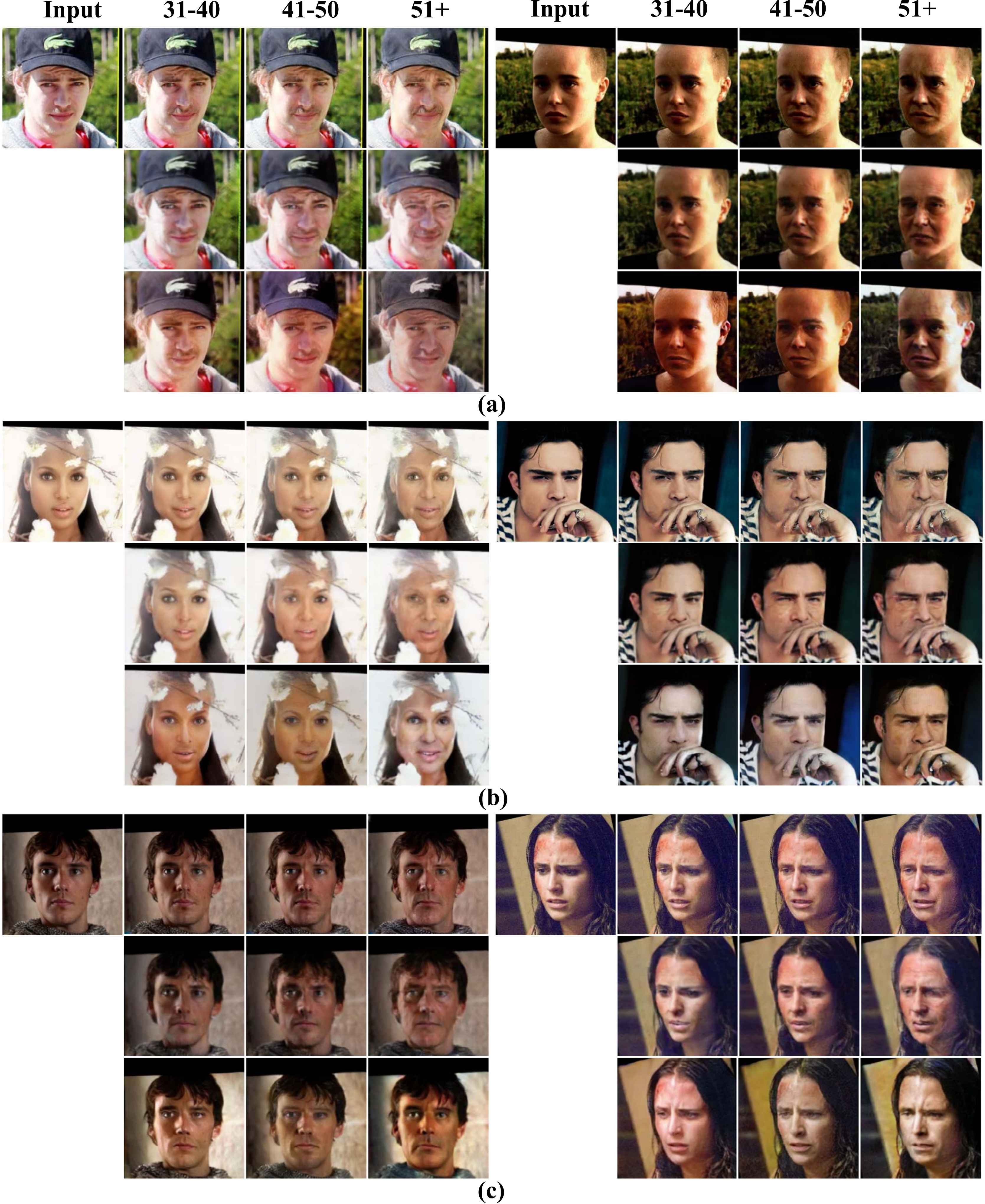}
    \caption{The generated aged faces under three extreme conditions of face on the CACD dataset:~(a) illumination,~(b) occlusion,~(c) low quality. Each input face was aged into three old age groups by our PFA-GAN~(first row), IPCGAN~(second row), and PAG-GAN~(third row).}
    \label{fig:robustness_analysis}
\end{figure}

The qualitative results above show that PFA-GAN is robust to various conditions such as pose, expression, and occlusion, during face aging and rejuvenation. For example, the occlusion, such as glasses and jewelry on the faces, is well preserved during face rejuvenation. Here, we further investigate the robustness of the face aging model under extreme uncontrolled conditions. Fig.~\ref{fig:robustness_analysis} showcases some aged faces under three representative extreme conditions---including illumination, occlusion, and low quality---demonstrating the strong robustness of PFA-GAN over IPCGAN and PAG-GAN.

We revisit previous works to better explain why our PFA-GAN outperforms others under these extreme conditions. Specifically, the cGANs-based methods such as IPCGAN~\cite{wang2018face} achieve face aging with a single generator while GANs-based methods such as PAG-GAN~\cite{yang2018learning} train several generators separately to learn each mapping from young faces to elder ones. Due to the intrinsic complexities of face aging, they cannot generate high-quality aged faces with smooth aging effects, especially when the age gap becomes large, making them not robust under extreme conditions. On the contrary, the generator of PFA-GAN consists of several sub-generators, and the output of one sub-generator is fed into the next sub-generator. Once one sub-generator amid the whole aging process produces ghosted faces, the following sub-generators can detect such anomalies and enforce that generator to produce satisfied faces through back-propagation. In a sense, the latter sub-generators provide an attention mechanism for earlier ones to age faces effectively. Therefore, PFA-GAN is capable of focusing on these areas of the face image relevant to face aging through only learning the residual images---aging effects.

\begin{figure}[t!]
    \centering
    \includegraphics[width=1.0\linewidth]{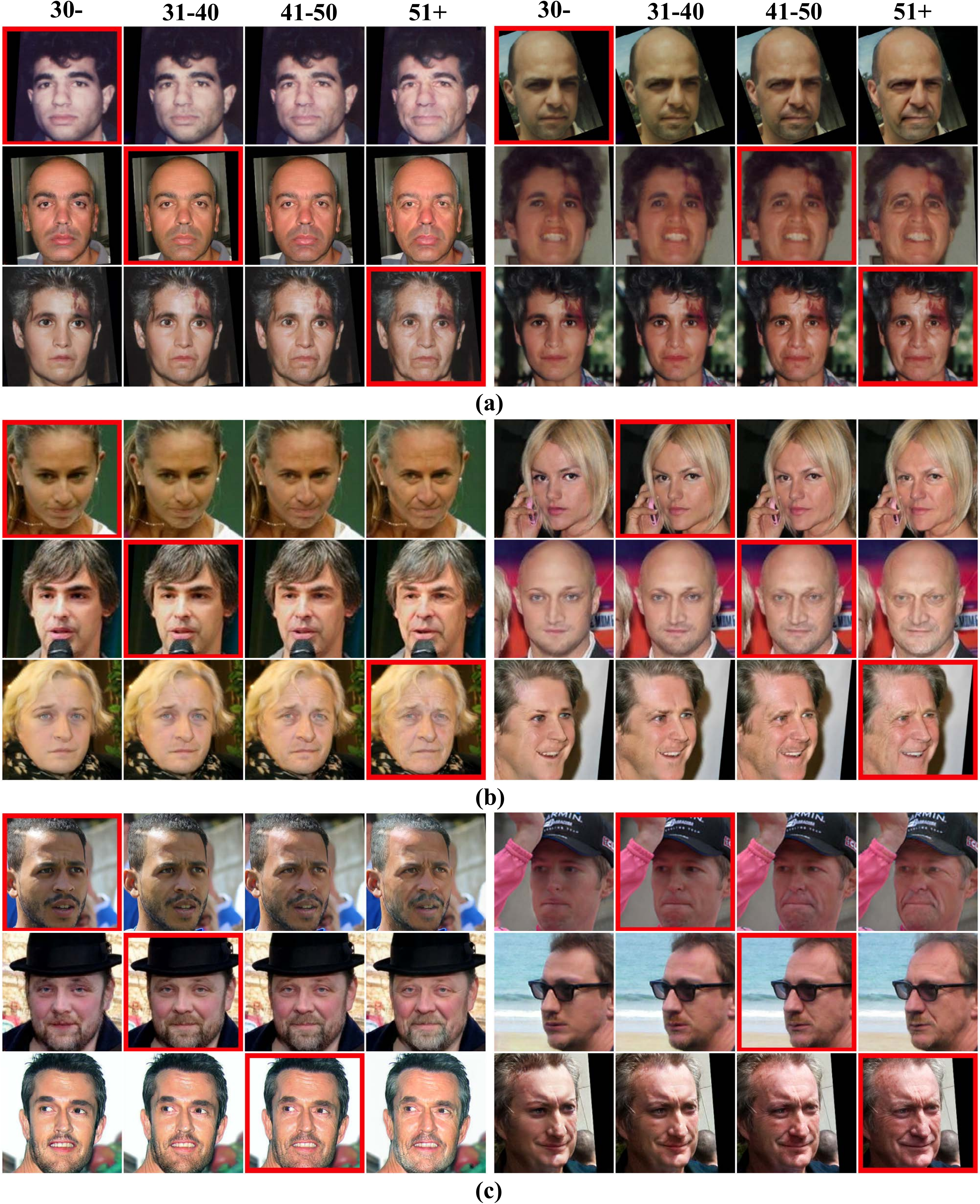}
    \caption{Sample results of both face aging and rejuvenation by applying our PFA-GAN model trained on the CACD dataset to three external datasets: (a) FG-NET~\cite{lanitis2002toward},~(b) CelebA~\cite{liu2015deep},~(c) IMDB-WIKI~\cite{Rothe-IJCV-2018}. Red boxes indicate input faces.}
    \label{fig:generalization_ability_study}
\end{figure}

\subsection{Generalization Ability}

To evaluate the generalization ability of PFA-GAN, we applied the model trained on the CACD dataset to external images from FG-NET~\cite{lanitis2002toward}, CelebA~\cite{liu2015deep}, and IMDB-WIKI~\cite{Rothe-IJCV-2018} datasets for face aging and rejuvenation. For those images without ground-truth age labels, face ages were estimated by the Face++ APIs. Fig.~\ref{fig:generalization_ability_study} presents exciting results, demonstrating that PFA-GAN generalizes well to face images with different sources for both face age progression and regression. Noticeably, the occlusions on the input faces, such as the makeup, scars, and glasses, are also well preserved in the aged faces.

\begin{figure*}[t]
    \centering
    \includegraphics[width=1\linewidth]{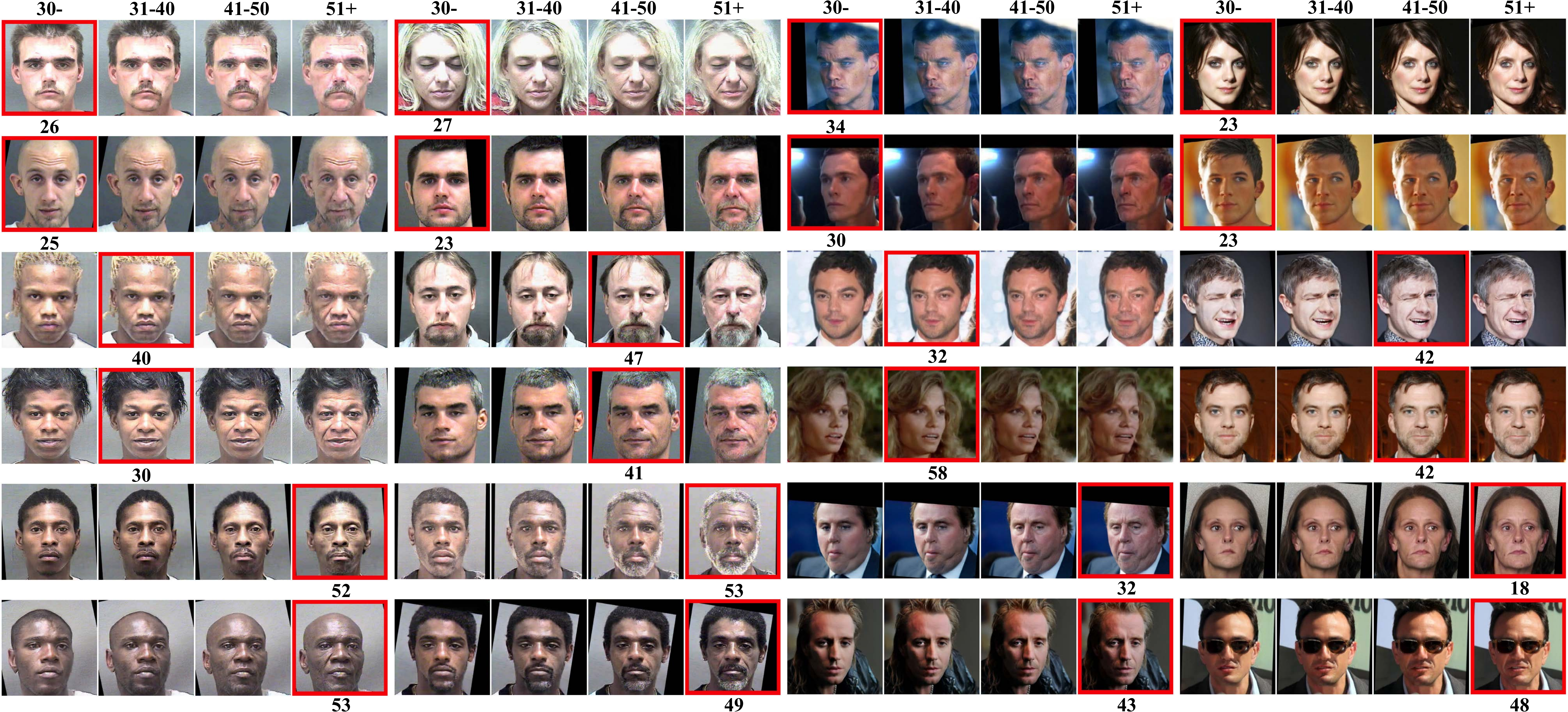}
    \caption{Sample results of PFA-GAN on the MORPH~(left two columns) and CACD~(right two columns) datasets for both face aging and rejuvenation. To examine the performance of PFA-GAN when the source age is unavailable, we estimate the age of a given face by the trained age estimation network $A$. Red boxes indicate input faces with the ground truth age label below it. Note that not all ages are estimated correctly.}
    \label{fig:limitation_study}
\end{figure*}

\subsection{Limitations}

Although PFA-GAN can achieve state-of-the-art performance both qualitatively and quantitatively, some limitations exist in PFA-GAN. First, compared to cGANs-based methods, the major limitation of the GANs-based methods including PFA-GAN, PAG-GAN~\cite{yang2018learning}, and A3GAN~\cite{liu2019attribute}, is that the networks require the source age label as the input to achieve the aging process. To eliminate the doubts towards the effectiveness of PFA-GAN, we removed the source age label in the testing phase and instead estimated it by the trained age estimation network $A$. Fig.~\ref{fig:limitation_study} shows the sample results for both face age progression and regression. Even though some faces are misclassified into other age groups, the results verify that our PFA-GAN is robust even with noisy age labels. Besides, considering that the faces age progressively, PFA-GAN can produce more smooth aging results than PAG-GAN~\cite{yang2018learning}.
Second, PFA-GAN needs to re-train another model for rejuvenation, although what we need to do is only to reverse the order of the age groups for the training of face rejuvenation. However, PFA-GAN can achieve face progression and rejuvenation in one model if it uses invertible neural network such as~\cite{kingma2018glow} as the sub-network. Since this paper mainly presents the progressive face aging framework for face aging, the invertible neural network for face progression and rejuvenation deserves studying as a future work. 

Third, with more age groups split in face aging, the difficulty of end-to-end training from scratch for our network would be increased and the patterns between two adjacent age groups will become less clear. The possible solutions are to first train each sub-network independently and then fine-tune these sub-networks in an end-to-end manner, and use more data to characterize the aging patterns between two adjacent age groups, respectively. 

Last, PFA-GAN may produce worse background compared to the one without progressive module as shown in Fig.~\ref{fig:ablation_study}. This may be caused by the simple background presented in the dataset and could be addressed with diverse background like CACD dataset.

%% file: conc.tex
\section{Conclusion}\label{sec:conc}

In this paper, we proposed a novel progressive face aging framework based on generative adversarial networks~(PFA-GAN) to model the age progression by a progressive neural network. In doing so, PFA-GAN aged the input young face in a progressive way to mimic the human face age progression. We also introduced a novel age estimation loss and an aging smoothness metric. The PFA-GAN can be optimized in an end-to-end manner to eliminate the accumulative error. Experimental results on two benchmarked datasets demonstrated the superiority of PFA-GAN over the state-of-the-art cGAN-based methods in terms of image quality, aging accuracy, aging smoothness, and identity preservation.

%% file: appendix.tex
\clearpage
\onecolumn
\appendix
\addcontentsline{toc}{section}{Appendix}
\numberwithin{equation}{section}
\setcounter{figure}{0}
\renewcommand\thefigure{\thesection.\arabic{figure}} 
\renewcommand\thetable{\thesection.\arabic{table}}

\begin{figure*}[htbp]
    \centering
    \includegraphics[width=0.6\linewidth]{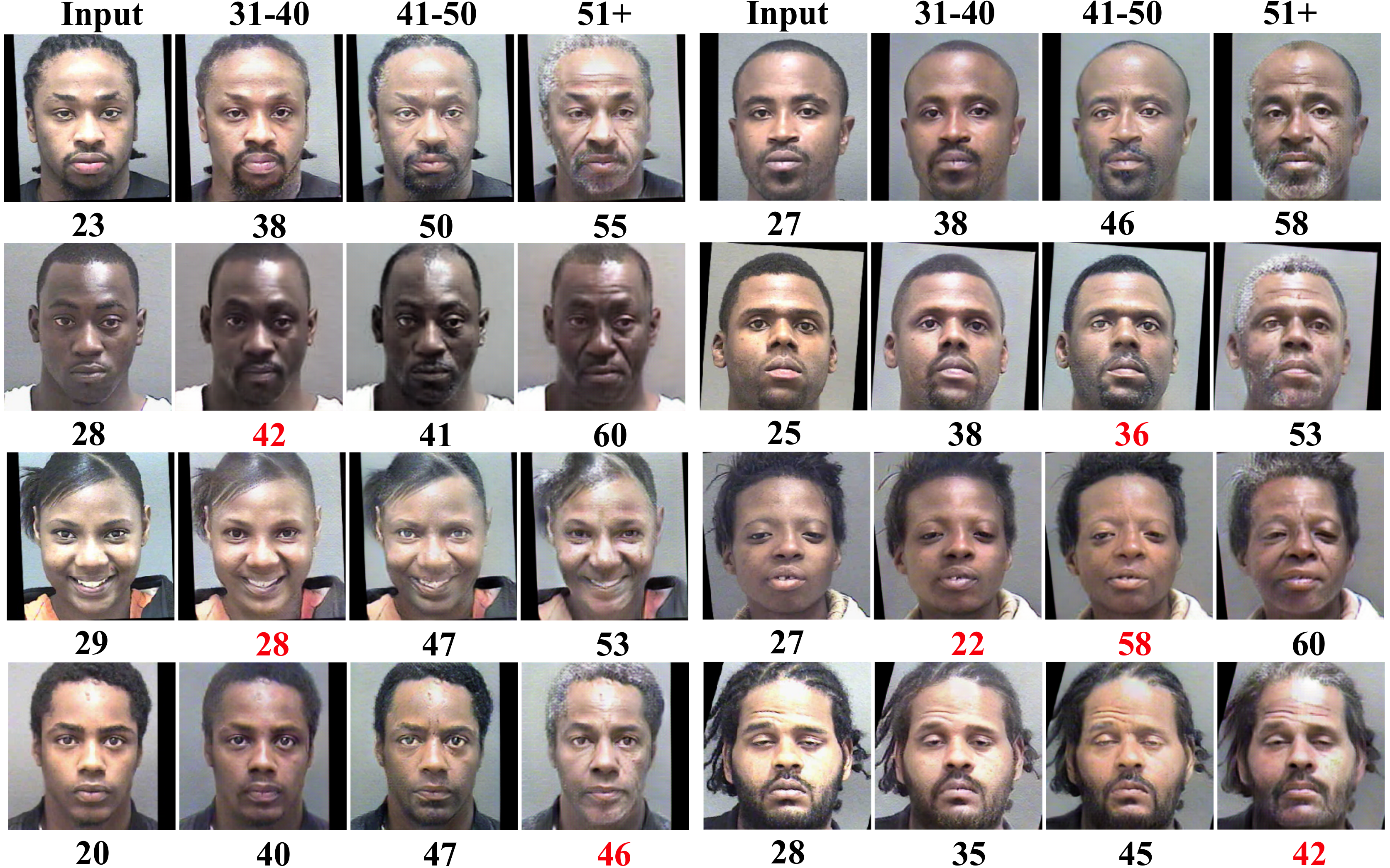}
    \caption{Examples of unsmoothed face aging generated by PAG-GAN~\cite{yang2018learning}. The young faces from the age group $30-$ are aged into the three old age groups, whose appearance ages are obtained via publicly available Face++ APIs~\cite{faceplusplus.com}. The estimated ages are placed below images, where the first row shows two expected normal samples and the resting rows shows some unsmoothed ages highlighted in red color.}
    \label{fig:not_smooth_vis}
\end{figure*}

\begin{figure*}[htbp]
    \centering
    \includegraphics[width=1.0\linewidth]{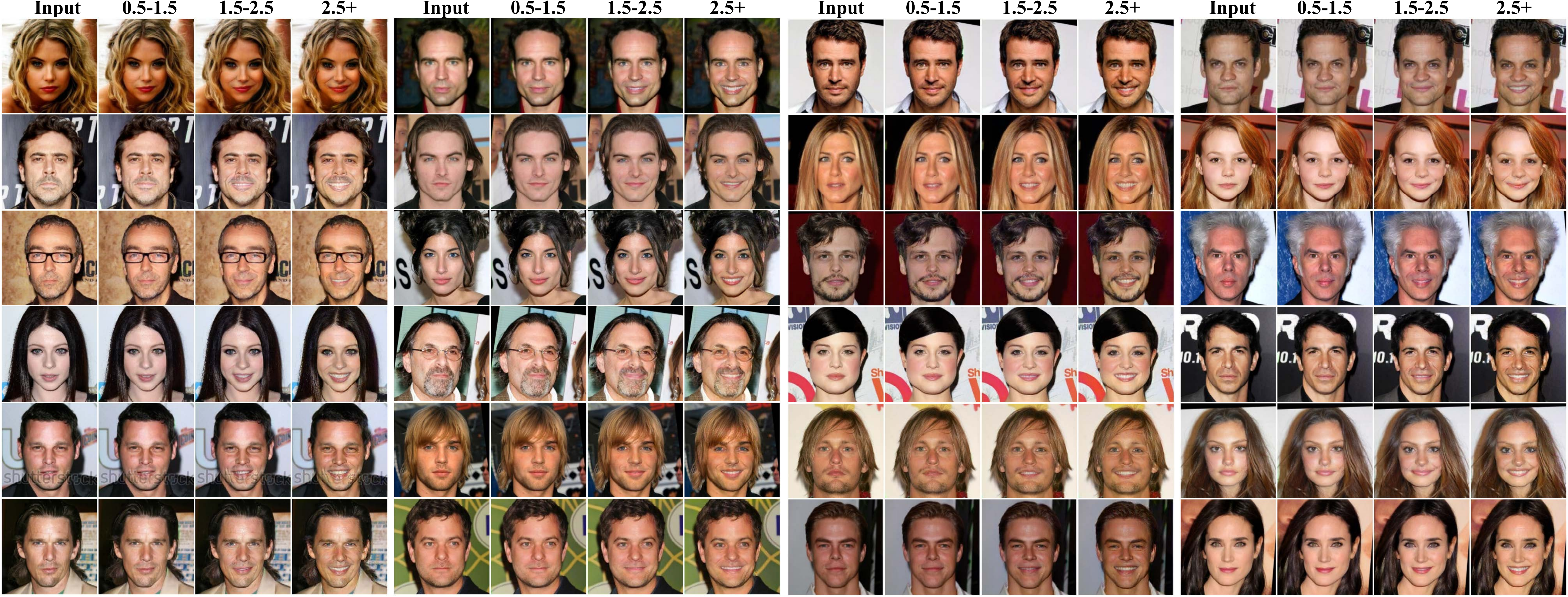}
    \caption{Sample results achieved by PFA-GAN on the CACD dataset for facial expression translation. First, following~\cite{Pumarola_2018_ECCV}, the CACD dataset was annotated with~\cite{tadas2015cross} to obtain continuous action units~(AUs)~\cite{Ekman1978FacialAC}. Second, we only utilize the AU12 for smile-like facial expression translation as the age label for face aging, and we roughly split the data into 4 groups according to the annotated AU12 intensity ranging from $0$ to $5$ for the sake of data balance; \ie, $0.5-$, $0.5-1.5$, $1.5-2.5$ and $2.5+$. At last, PFA-GAN was adapted to achieve facial expression translation and the testing faces came from $0.5-$ group.}
    \label{fig:expression_translation}
\end{figure*}

\begin{table*}[ht]
    \centering
    \caption{Detailed network architecture of PFA-GAN.}
    \label{tab:network_architecture}
    \begin{tabular}{|p{0.10\textwidth}|c|c|c|ccccc|}
        \hline
        \multicolumn{9}{|c|}{\textbf{Sub-network}}\\ \hline
        {Layer} & Filter/Stride & Normalization & Activation & \multicolumn{5}{c|}{Output Shape}     \\ \hline
        Conv                  & $9\times 9/1$ & Intance       & LReLU      & 32  & $\times$ & 256 & $\times$ & 256 \\ \hline
        Conv                  & $4\times 4/2$ & Intance       & LReLU      & 64  & $\times$ & 128 & $\times$ & 128 \\ \hline
        Conv                  & $4\times 4/2$ & Intance       & LReLU      & 128 & $\times$ & 64  & $\times$ & 64  \\ \hline
        ResBlock$\times 4$    & $3\times 3/1$ & Intance       & LReLU      & 128 & $\times$ & 64  & $\times$ & 64  \\ \hline
        Deconv                & $4\times 4/2$ & Intance       & LReLU      & 64  & $\times$ & 128 & $\times$ & 128 \\ \hline
        Deconv                & $4\times 4/2$ & Intance       & LReLU      & 32  & $\times$ & 256 & $\times$ & 256 \\ \hline
        Conv                  & $9\times 9/1$ & -             & -          & 3   & $\times$ & 256 & $\times$ & 256 \\ \hline
        \end{tabular}
    \begin{tabular}{|p{0.10\textwidth}|c|c|c|ccccc|}
        
        \multicolumn{9}{|c|}{\textbf{Discriminator}}\\ \hline
        {Layer} & Filter/Stride & Normalization & Activation & \multicolumn{5}{c|}{Output Shape}     \\ \hline
        Conv                   & $4\times 4/2$ & -             & LReLU      & 64  & $\times$ & 128 & $\times$ & 128 \\ \hline
        Conv                   & $4\times 4/2$ & Spectral      & LReLU      & 128 & $\times$ & 64  & $\times$ & 64  \\ \hline
        Conv                   & $4\times 4/2$ & Spectral      & LReLU      & 256 & $\times$ & 32  & $\times$ & 32  \\ \hline
        Conv                   & $4\times 4/2$ & Spectral      & LReLU      & 512 & $\times$ & 16  & $\times$ & 16  \\ \hline
        Conv                   & $4\times 4/1$ & Spectral      & LReLU      & 512 & $\times$ & 15  & $\times$ & 15  \\ \hline
        Conv                   & $4\times 4/1$ & -             & -          & 1 & $\times$ & 14  & $\times$ & 14  \\ \hline
    \end{tabular}
    \begin{tabular}{|p{0.10\textwidth}|c|c|c|ccccc|}
        
        \multicolumn{9}{|c|}{\textbf{Age Estimation Network}}\\ \hline
        {Layer} & Filter/Stride & Normalization & Activation & \multicolumn{5}{c|}{Output Shape}     \\ \hline
        Conv                    & $4\times 4/2$ & Batch         & ReLU       & 64  & $\times$ & 128 & $\times$ & 128 \\ \hline
        Conv                    & $4\times 4/2$ & Batch         & ReLU       & 128 & $\times$ & 64  & $\times$ & 64  \\ \hline
        Conv                    & $4\times 4/2$ & Batch         & ReLU       & 256 & $\times$ & 32  & $\times$ & 32  \\ \hline
        Conv                    & $4\times 4/2$ & Batch         & ReLU       & 512 & $\times$ & 16  & $\times$ & 16  \\ \hline
        Conv                    & $4\times 4/2$ & Batch         & ReLU       & 512 & $\times$ & 8   & $\times$ & 8   \\ \hline
        Conv                    & $4\times 4/2$ & Batch         & ReLU       & 512 & $\times$ & 4   & $\times$ & 4   \\ \hline
        Linear                  & $101$         & -             & -          &     &          & 101 &          &     \\ \hline
        Linear                  & $N$           & -             & -          &     &          & $N$ &          &     \\ \hline
    \end{tabular}
    
\end{table*}